\documentclass[sigconf]{acmart}
\AtBeginDocument{%
  }

\copyrightyear{2026}
\acmYear{2026}
\setcopyright{cc}
\setcctype{by}
\acmConference[KDD '26]{Proceedings of the 32nd ACM SIGKDD Conference on Knowledge Discovery and Data Mining V.1}{August 09--13, 2026}{Jeju Island, Republic of Korea}
\acmBooktitle{Proceedings of the 32nd ACM SIGKDD Conference on Knowledge Discovery and Data Mining V.1 (KDD '26), August 09--13, 2026, Jeju Island, Republic of Korea}
\acmPrice{}
\acmDOI{10.1145/3770854.3780214}
\acmISBN{979-8-4007-2258-5/2026/08}

\setcopyright{none}
\renewcommand\footnotetextcopyrightpermission[1]{}
\usepackage{newfloat}
\usepackage{listings}
\usepackage{algorithm}
\usepackage{adjustbox}
\usepackage{color}
\usepackage{xcolor}
\usepackage{colortbl}
\usepackage{array}
\usepackage{multirow}
\usepackage{booktabs}
\usepackage{pifont}
\usepackage{amsmath,amsfonts}
\usepackage{enumitem}
\usepackage[most]{tcolorbox}
\usepackage{float}
\usepackage{makecell}
\usepackage{subcaption}
\usepackage{xspace}
\usepackage{cleveref}
\usepackage{hyperref}
\usepackage{xcolor}
\usepackage{tcolorbox}
\usepackage{enumitem}
\usepackage{multicol}
\usepackage{algpseudocode}
\usepackage{cuted}
\tcbuselibrary{breakable}

\hypersetup{
    hypertex=true,
    colorlinks=true,
    linkcolor=black,
    anchorcolor=black,
    citecolor=black
}

\newcommand{\M}{TAME\xspace}
\newcommand{\RAG}{RA\textsuperscript{2}G\xspace}
\newcommand{\B}{LCMP\xspace}
\newcommand{\A}{personalized MLLM assistant\xspace}
\newcommand{\task}{MLLM Personalization\xspace}

\newcommand{\bc}[1]{\textcolor{blue}{#1}}

\usepackage{arydshln}
\usepackage{amsthm}

\crefname{figure}{Fig.}{Figs.}

\newtheoremstyle{customremark}
  {0.3em}
  {0.3em}
  {\itshape}
  {}
  {\bfseries}
  {.}
  {0.3em}
  {\thmname{#1}~\thmnumber{#2}}

\theoremstyle{customremark}
\newtheorem{remark}{Remark}

\newtheoremstyle{customdefinition}
  {0.3em}
  {0.3em}
  {\itshape}
  {}
  {\bfseries}
  {.}
  {0.3em}
  {\thmname{#1}~\thmnumber{#2}\thmnote{~(#3)}}

\theoremstyle{customdefinition}
\newtheorem{definition}{Definition}

\newlength\savewidth
\newcommand\shline{\noalign{\global\savewidth\arrayrulewidth
                            \global\arrayrulewidth 1pt}%
                   \hline
                   \noalign{\global\arrayrulewidth\savewidth}}

\NewDocumentCommand{\rp}{O{comment} m}{%
  \IfEqCase{#1}{%
    {comment}{\textcolor{red}{\textbf{[@RP COMMENT: #2]}}}%
    {done}{\textcolor{green}{\textbf{@RP DONE: [#2]}}}%
    {todo}{\textcolor{blue}{\textbf{@RP TODO: [#2]}}}%
  }[\textcolor{gray}{#2}]%
}

\tcbuselibrary{listings,breakable,skins}

\definecolor{promptblue}{RGB}{4,90,141}
\definecolor{lightblue}{RGB}{241,238,246}

\newtcolorbox{promptbox}[1]{
    colback=lightblue,
    colframe=promptblue,
    arc=8pt,
    boxrule=2pt,
    title={\Large\textbf{#1}},
    coltitle=white,
    colbacktitle=promptblue,
    fonttitle=\bfseries,
    titlerule=0pt,
    toptitle=8pt,
    bottomtitle=8pt,
    left=10pt,
    right=10pt,
    top=6pt,
    bottom=10pt,
    breakable,
    fontupper=\ttfamily
}

\NewDocumentEnvironment{promptbox*}{m}{%
    \begin{tcolorbox}[
        colback=lightblue,
        colframe=promptblue,
        arc=8pt,
        boxrule=2pt,
        title={\Large\textbf{#1}},
        coltitle=white,
        colbacktitle=promptblue,
        fonttitle=\bfseries,
        titlerule=0pt,
        toptitle=4pt,
        bottomtitle=4pt,
        left=10pt,
        right=10pt,
        top=10pt,
        bottom=10pt,
        width=\textwidth,
        fontupper=\ttfamily
    ]
}{%
    \end{tcolorbox}
}

\begin{document}

\title[Towards Training-Free and State-Aware Personalized MLLM Assistant]{TAMEing Long Contexts in Personalization: Towards Training-Free and State-Aware MLLM Personalized Assistant}

\author{Rongpei Hong}
\email{rongpei.hong@std.uestc.edu.cn}
\orcid{0009-0007-4977-1657}
\affiliation{%
  \institution{University of Electronic Science and Technology of China}
  \city{Chengdu}
  \state{Sichuan}
  \country{China}
}

\author{Jian Lang}
\email{jian_lang@std.uestc.edu.cn}
\orcid{0009-0009-0876-0497}
\affiliation{%
  \institution{University of Electronic Science and Technology of China}
  \city{Chengdu}
  \state{Sichuan}
  \country{China}
}

\author{Ting Zhong}
\authornote{Corresponding author.}
\email{zhongting@uestc.edu.cn}
\orcid{0000-0002-8163-3146}
\affiliation{%
  \institution{University of Electronic Science and Technology of China}
  \city{Chengdu}
  \state{Sichuan}
  \country{China}}

\author{Yong Wang}
\orcid{0000-0002-8699-8355}
\email{wangyong@ipplus360.com}
\affiliation{%
  \institution{Aiwen Technology Co., Ltd.}
  \city{Zhengzhou}
  \state{Henan}
  \country{China}}

\author{Fan Zhou}
\email{fan.zhou@uestc.edu.cn}
\orcid{0000-0002-8038-8150}
\affiliation{%
  \institution{University of Electronic Science and Technology of China}
  \city{Chengdu}
  \state{Sichuan}
  \country{China}
}
\affiliation{%
  \institution{Intelligent Digital Media Technology Key Laboratory of Sichuan Province}
  \city{Chengdu}
  \state{Sichuan}
  \country{China}
}

\renewcommand{\shortauthors}{Rongpei Hong et al.}

\begin{abstract}
Multimodal Large Language Model (MLLM) Personalization is a critical research problem that facilitates personalized dialogues with MLLMs targeting specific entities (known as personalized concepts).
However, existing methods and benchmarks focus on the simple, \textit{context-agnostic} visual identification and textual replacement of the personalized concept  (e.g., ``A yellow puppy'' \textrightarrow~``Your puppy Mochi''), overlooking the ability to support long-context conversations.
An \textit{ideal} \A is capable of engaging in \textit{long-context} dialogues with humans and continually improving its experience quality by learning from past dialogue histories.
To bridge this gap, we propose \textbf{\B}, the first Long-Context MLLM Personalization evaluation benchmark.
\B assesses the capability of MLLMs in perceiving variations of personalized concepts and generating contextually appropriate personalized responses that reflect these variations.
As a strong baseline for \B, we introduce a novel training-free and state-aware framework \textbf{\M}.
\M endows MLLMs with double memories to manage the temporal and persistent variations of each personalized concept in a differentiated manner.
In addition, \M incorporates a new training-free Retrieve-then-Align Augmented Generation (\RAG) paradigm.
\RAG introduces an alignment step to extract the contextually fitted information from the multi-memory retrieved knowledge to the current questions, enabling better interactions for complex real-world user queries.
Experiments on \B demonstrate that \M achieves the best performance, showcasing remarkable and evolving interaction experiences in long-context scenarios.
\end{abstract}



\maketitle

\section{Introduction}
\label{sec:intro}
In the not-so-distant future, when you talk to your AI assistant, it can truly \textit{remember} you.
Imagine that in such a moment, you ask the assistant: ``What should my pet kitten eat this morning?''.
The assistant analyzes a live photo of your kitten, recognizing signs of physical fatigue and loss of appetite. 
It then revisits a previous dialogue where you mentioned that your kitten was recently diagnosed with chronic kidney disease.
Combining these visual cues with contextual information, it responds: ``Since your kitten looks tired this morning and has chronic kidney disease, a warmed-up portion of low-phosphorus prescription food would be ideal.''
This scenario, which combines \textit{personalized} understanding and generation with \textit{visual} and \textit{textual} information, is closely tied to the core task of Multimodal Large Language Model (MLLM) Personalization~\cite{alaluf2024myvlm, nguyen2024yollava}.

\begin{figure}[t]
  \centering
  {\includegraphics[width=0.97\columnwidth]{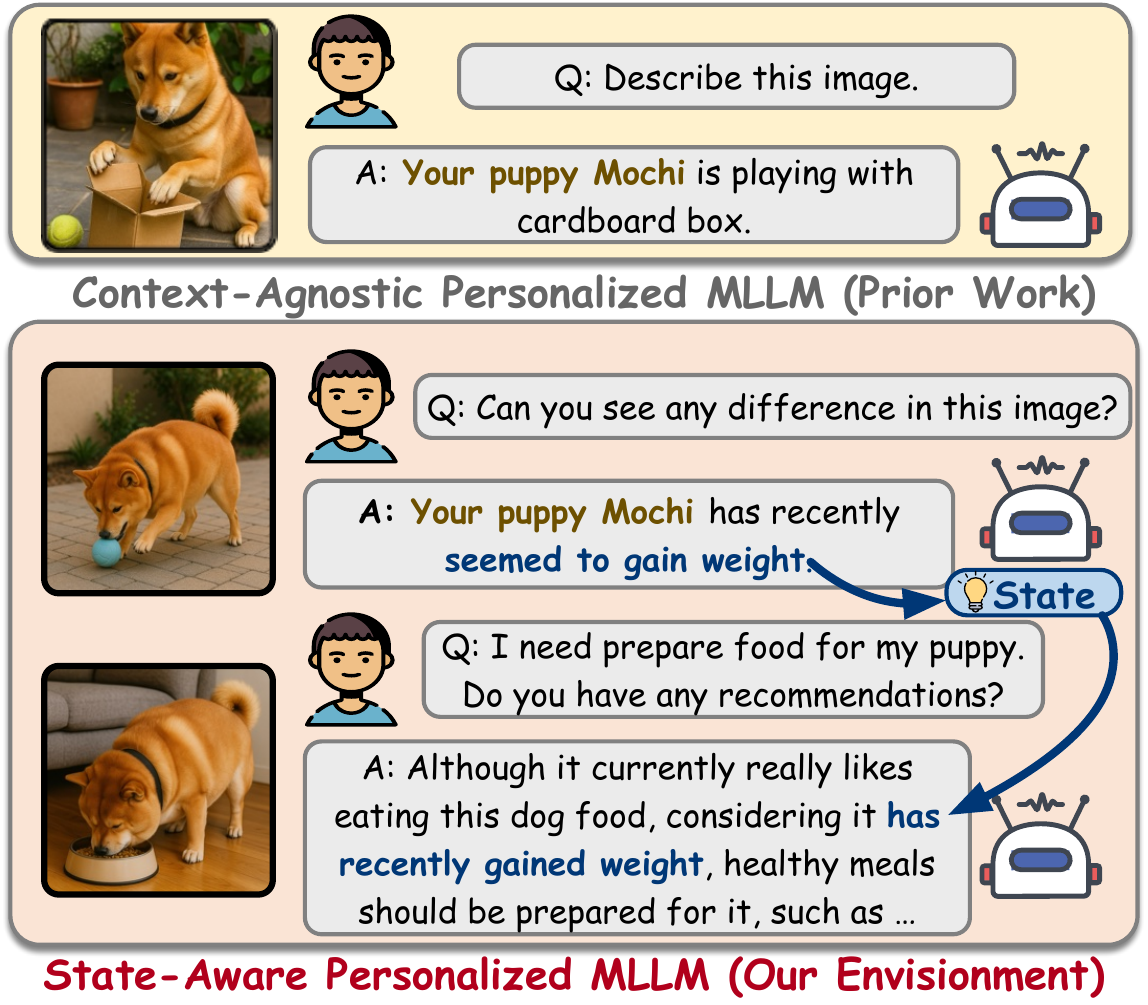}}
  \vspace{-3mm}
  \caption{Comparison with prior work: Our envisioned state-aware personalized MLLM assistant learns from historical dialogue to improve the quality of ongoing conversations.}
  \label{fig:intro}
  \vspace{-4mm}
\end{figure}

An important notion in \task is the \textit{personalized concept}, which can be defined as a user-specific entity rather than a generic one (e.g., your dog Charlie, my best friend Calvin), characterized by its unique attributes (e.g., name, appearance, preference, and connection).
As illustrated in~\Cref{fig:intro}, existing works in this field focused on improving the MLLM's capability in simple visual identification and textual replacement of personalized concepts~\cite{alaluf2024myvlm, nguyen2024yollava, hao2024rap, pi2025personalized}, such as rephrasing an image caption from ``A yellow puppy is playing with a cardboard box'' to ``Your puppy Mochi is playing with a cardboard box.''
Among these works, MyVLM~\cite{alaluf2024myvlm} added external classifiers to recognize specific concepts and learned an embedding for each concept to personalize the outputs of MLLMs.

\textit{Although} significant progress has been made toward the current goal of \task, existing works and evaluation benchmarks remain single-turn and \textit{context-agnostic}, as MLLMs neither maintain dialogue state nor leverage contextual information for more coherent and high-quality interactions with the user.
\textit{In contrast}, we envision a \A that learns from past interaction contexts to continually improve the interactive experience with the user.
In this work, we concretize this improvement in interactive experience during prolonged interactions as follows: \textit{a \A can refine the modeling of attributes to a personalized concept based on past dialogues, enabling more accurate and contextually appropriate personalized responses when referring to this concept in future conversations}.
For instance, as shown in \Cref{fig:intro}, an intelligent \A should learn from the historical dialogue context to recognize that your puppy Mochi has become fatter recently (a change in its physical condition attribute) and should therefore suggest a light diet.

In this work, we focus on perceiving and modeling the variations of \textit{long-term} and \textit{short-term} attributes of personalized concepts during prolonged interactions for improved interactive experience in \task.
The long-term attributes refer to relatively stable properties that persist with the personalized concept (e.g., pet’s name, favorite food). 
In contrast, short-term attributes refer to transient and frequently updated properties that are effective only within the immediate short-term context following their modification (e.g., ongoing activity, recent physical condition).
To this end, we propose the first \textbf{\underline{L}}ong-\textbf{\underline{C}}ontext \textbf{\underline{M}}LLM \textbf{\underline{P}}ersonalization evaluation benchmark (\textbf{\B}).
Building upon a well-designed automatic pipeline, \B endows each personalized concept with fine-grained temporal-scale attributes across both short- and long-term horizons, and constructs multi-turn vision-language dialogues that incorporate diverse user-driven modifications over these temporal properties.
The evaluation is designed to assess the capability of an MLLM to manage the variations of both types of attributes by combining and interpreting them within extended dialogues, thereby generating contextually fitted personalized responses.

To cope with the \task under long context scenarios, we propose \textbf{\M}, a novel \underline{\textbf{T}}raining-free and state-\underline{\textbf{A}}ware Personalized MLLM Assistant powered by Double \underline{\textbf{ME}}mories.
As a strong baseline for \B, our framework equips the \A with a \textit{double-memory architecture}: one memory stores \textit{static} and universal personalized knowledge, while the other monitors \textit{dynamic} contextual information throughout the interaction.
Specifically, we first introduce a fresh \textit{Dynamic State Memory} (DS Memory) to continually track multimodal contextual information throughout the dialogue, with a focus on capturing any variations to the short- and long-term attributes of personalized concepts.
In addition, we propose a new \textit{Static Personalized Memory} (SP Memory), which maintains each personalized concept along with its associated long-term and stable attributes.
To ensure the efficiency and timeliness of the DS Memory, we introduce a novel \textit{Double Memory Transition} mechanism, which updates the long-term attributes in the SP Memory by referencing the DS Memory, while discarding outdated short-term attributes from earlier dialogue stages.
This mechanism enables the MLLM to gradually become more \textit{intelligent} and better \textit{aligned} with the user through continuous interactions.
Finally, unlike prior works that rely on costly training to make MLLMs ``remember'' abundant personalized concepts~\cite{alaluf2024myvlm, nguyen2024yollava, pi2025personalized}, we draw inspiration from retrieval-augmented generation (RAG) techniques~\cite{lewis2020retrieval} in modern LLMs~\cite{brown2020language, touvron2023llama} and introduce a novel training-free \textit{Retrieve-then-Align Augmented Generation} (\textbf{\RAG}) recipe.
Concretely, \RAG first retrieves question-related personalized information from both static and dynamic memories.
It then organizes and aligns the multi-memory retrieved content with the user’s query to generate responses grounded in the most relevant while appropriate multimodal contextual information.
Our main contributions are listed as follows:
\begin{itemize}[leftmargin=*,
                topsep=0pt plus 0.5pt,
                partopsep=0pt plus 0.5pt,
                itemsep=0pt plus 0.5pt,
                parsep=0pt plus 0.5pt]
\item To the best of our knowledge, we pioneer unearthing long-context scenarios as a more valuable direction for \task.
We then propose \textbf{\B}, the first benchmark for evaluating Long-Context MLLM Personalization. 
\item We introduce \textbf{\M}, a novel training-free and state-aware \A powered by double memories.
As a strong baseline for \B, \M is capable of perceiving fine-grained variations in both short- and long-term attributes of personalized concepts, and generating responses that are faithful to appropriate multimodal contextual information.
\item Extensive experiments and cases conducted on \B showcase that \M achieves remarkable personalized interaction quality in long-context vision-language conversations, establishing a strong baseline for long-context \task.

\end{itemize}

The code and data for our proposed \B and \M are available at \bc{\url{https://github.com/ronpay/TAME}}.

\section{Related Work}
\label{sec:related}

\subsection{MLLM Personalization}
\task aims at enabling MLLM to remember and understand user-specific concepts for personalized language response generation~\cite{alaluf2024myvlm, nguyen2024yollava, hao2024rap}.
Pioneering work MyVLM~\cite{alaluf2024myvlm} and Yo'LLaVA~\cite{nguyen2024yollava} brought initial definition and objective for this task with effective methods proposed.
MyVLM introduced external classification heads to recognize specific concepts and learned an
embedding for each concept for personalized generation of MLLMs.
Yo'LLaVA learned specific prompt tokens for the identification and personalized generation of each concept.
However, their approaches require frequent test-time training for each newly introduced personalized concept, which leads them less practical for real-world scenarios.
Following studies attempted to alleviate the issue of repetitive training by conducting prior large-scale fine-tuning on MLLM~\cite{hao2024rap, pi2025personalized}.
For instance, PVIT~\cite{pi2025personalized} employed visual fine-tuning on large-scale personalized dataset to equip MLLMs with the capability for personalized conversations.
Nevertheless, both repetitive test-time training and large-scale fine-tuning are computationally expensive and not applicable to closed-source MLLMs (e.g., GPT-4o~\cite{achiam2023gpt}), which significantly limits their practical deployment.
Furthermore, current goal of \task has primarily focused on simple visual recognition and textual substitution of personalized concepts.
Consequently, existing methods remain context-agnostic and thus fail to improve the quality of the interactive experience during prolonged human interactions.

To bridge this gap, we propose the \textbf{\B} benchmark, which evaluates the effectiveness of a \A in leveraging dialogue context to generate more contextually fitted responses.
To adapt \B, we introduce a novel training-free and state-aware baseline method \textbf{\M}.
\M maintains double memories to store both static, universal personalized concept information and dynamic contextual information.
It then designs a new \RAG paradigm, which effectively harnesses these memories to generate more contextually appropriate personalized responses in a training-free manner.
Concurrent work PeKit~\cite{seifi2025personalization} and R2P~\cite{das2025training} also realized training-free \task.
However, they remain stateless and fail to progressively improve the experience quality of the MLLM assistant over prolonged user interactions.

\subsection{Retrieval-Augmented Generation}
Retrieval-augmented generation (RAG) has emerged as a powerful technique for enhancing the generative capabilities of large language models (LLMs) by integrating external knowledge retrieval, rather than relying solely on pre-trained parametric knowledge~\cite{lewis2020retrieval,  fan2024survey,che2024hierarchical, wu2024coral, jiang2025piperag}.
RAG offers greater scalability and enables LLMs to adapt to new datasets and scenarios without requiring parameter updates, making it a more cost-effective and flexible solution.
For instance, Self-Rag~\cite{asai2023self} employed special tokens to make retrieval adaptive and controllable. 
PipeRAG~\cite{jiang2025piperag} enables low-latency RAG by pipelining retrieval and generation with flexible intervals and a system-aware performance model.
Recent work also explored introducing RAG into MLLMs~\cite{long2025retrieval, suri2025visdom, lang2025retrievalaugmented, hou2025fire}.
For example, ReAuSE~\cite{long2025retrieval} replaces discriminative retrieval with autoregressive document ID generation, enabling seamless integration of retrieval into MLLMs.
Inspired by the cost-efficiency of RAG, we introduce a novel \RAG paradigm for long-context \task.
\RAG is specifically designed for our \M, enabling it to coordinate knowledge from multiple memories and align the most relevant contextual information for personalized generation.

\begin{table}[t]
\centering
\caption{Comparison among prior benchmarks in \task and our proposed \B.}
\vspace{-3mm}
\label{tab:benchmark_compare}
\setlength{\tabcolsep}{3.2pt}
\begin{adjustbox}{width=\linewidth}
\begin{tabular}{lccccc}
\toprule
& \multicolumn{2}{c}{\textbf{Personalized Concept}} & \multicolumn{2}{c}{\textbf{Evaluation}} \\ \cmidrule(lr){2-3} \cmidrule(lr){4-5}
\textbf{Benchmark} & \textbf{Image} & \textbf{Attributes} & \textbf{Type} & \textbf{Context} \\ \midrule
MyVLM~\cite{alaluf2024myvlm} & \ding{51} & \ding{55} & Caption & \ding{55} \\
Yo'LLaVA~\cite{nguyen2024yollava} & \ding{51} & \ding{55} & Caption, VQA & \ding{55}\\
P-Bench~\cite{pi2025personalized} & \ding{51} & \ding{51} & Caption, VQA & \ding{55}\\ 
PerVA~\cite{das2025training} & \ding{51} & \ding{55} & Caption, VQA & \ding{55}\\
\midrule
\textbf{\B} & \ding{51} & \ding{51} (long, short-term) & Multi-Turn VQA & \ding{51}\\
\bottomrule
\end{tabular}
\end{adjustbox}
\vspace{-3mm}
\end{table}

\section{The Proposed \B Benchmark}
\label{sec:benchmark}
\subsection{Limitations of Prior Benchmarks}
In the context of \task, the notion of \textit{personalized concept} plays a central and important role.
\begin{definition}[Personalized Concept]
A personalized concept refers to a user-specific entity (e.g., ``my dog'') with an image as visual identifier and individualized attributes and features (e.g., ``fur color of my dog''), distinguishing it from generic concepts~\cite{alaluf2024myvlm, nguyen2024yollava}.
\end{definition}
Prior work in \task primarily focused on identifying the visual characteristics of personalized concepts and replacing generic objects with personalized references during generation~\cite{alaluf2024myvlm, nguyen2024yollava, pi2025personalized}.
\textit{Nevertheless}, such personalized MLLM assistants are not intelligent enough, overlooking the improvement of experience quality by learning from contextual information during prolonged interactions with humans.
To concretize such experience improvement, we propose the following plausible measurement:
\begin{remark}
A state-aware \A should be able to capture any changes in personalized concepts during real-world human interactions, specifically changes in the attributes associated with each concept, and generate contextually appropriate personalized responses when referring to such concepts in future interactions.
\end{remark}
\textit{However}, as shown in~\Cref{tab:benchmark_compare}, prior benchmarks either provide only \textit{coarse-grained} characterizations of personalized concepts (e.g., images and personalized captions only~\cite{alaluf2024myvlm, nguyen2024yollava}) or evaluate the capabilities of \A using \textit{context-agnostic} simplified tasks such as captioning or single-turn VQA (e.g., ``Is my car blue?'').
As a result, they fail to assess the ability of \A in leveraging contextual information from previous interactions to improve the quality of future interactions.

\begin{figure}[t]
  \centering
  {\includegraphics[width=0.98\columnwidth]{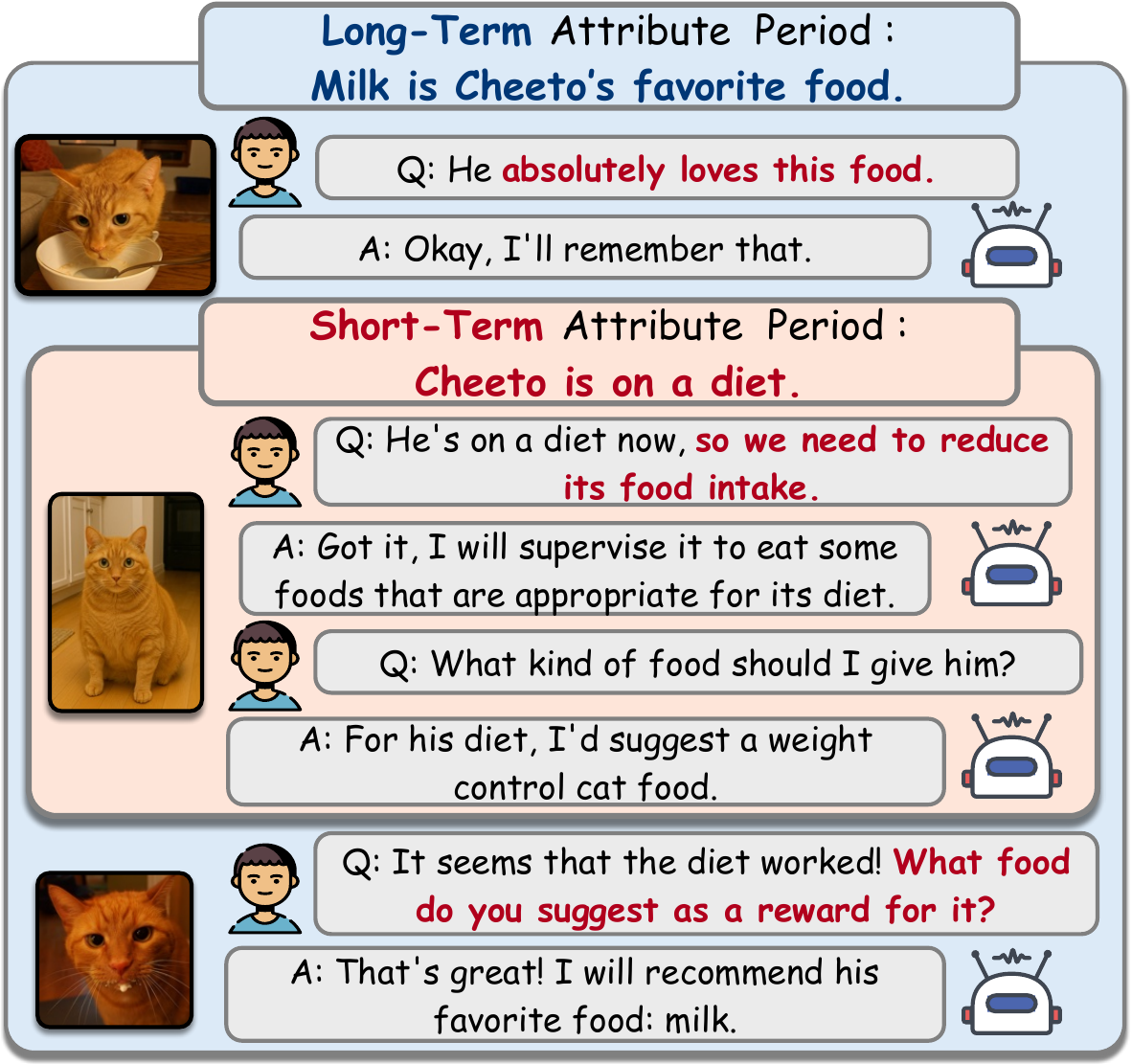}}
  \vspace{-3mm}
  \caption{Conceptual illustration of long-term (persistent) and short-term (temporal) attributes. Short-term attributes can override long-term ones to reflect dynamic context.}
  \label{fig:ls-attribute}
  \vspace{-3.5mm}
\end{figure}

\subsection{\B Benchmark Construction}
\label{subsec:benchmark-const}
In this study, to overcome the limitations of existing benchmarks in \task, we propose \textbf{\B}, the first long-context MLLM personalization evaluation benchmark.
As shown in~\Cref{tab:benchmark_compare}, our \B features context-aware evaluations through multi-turn VQA tasks and focuses on assessing the ability of \A to learn from past vision-language dialogues to capture variations in personalized concept attributes.
To further refine the evaluation, we define two types of attributes with different temporal scales observed during long-term human interactions:
\begin{definition}[Short-Term Attribute]
A short-term attribute is a temporal, transient, and frequently updated property of a personalized concept (e.g., recent physical condition of my pet dog).
As illustrated in~\Cref{fig:ls-attribute}, such an attribute is only valid within a short-term context following its most recent definition or modification.
\end{definition}

\begin{figure*}[t]
  \centering
  \includegraphics[width=0.97\linewidth]{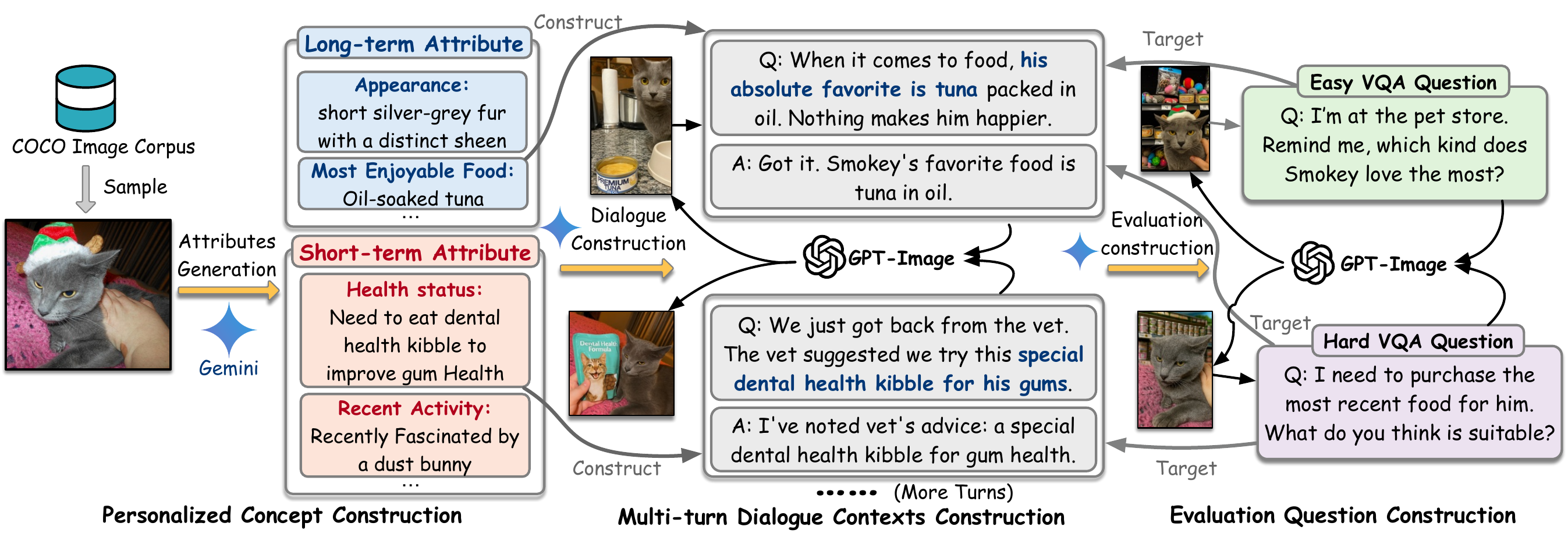}
    \vspace{-4mm}
  \caption{Construction pipeline of \B benchmark.}
  \label{fig:pipeline-const}
  \vspace{-3mm}
\end{figure*}

\begin{definition}[Long-Term Attribute]
A long-term attribute is a stable and persistent property of a personalized concept (e.g., the color of my car).
As illustrated in~\Cref{fig:ls-attribute}, such an attribute remains valid across long-term interactions, unless it is explicitly modified.
\end{definition}
Based on the definitions, as presented in~\Cref{fig:pipeline-const}, we present the automatic construction of \B from three progressive steps: (1) personalized concept construction, (2) multi-turn vision-language dialogue contexts construction, and (3) evaluation question construction.
For each \textbf{personalized concept construction}, we first randomly sample a raw image $\mathcal{I}$ that contains an entity (e.g., a cat) from open-source dataset Microsoft COCO~\cite{lin2014microsoft} to form a new personalized concept $\mathcal{C}$.
We then prompt an MLLM (Gemini-2.5-Pro~\cite{gemini2.5pro}) to endow the concept $\mathcal{C}$ with a set of long-term attributes $\mathcal{L} = \{L_i\}_{i=1}^{{O}_{L}}$ and short-term attributes $\mathcal{S} = \{S_i\}_{i=1}^{{O}_{S}}$, based on both its visual features and the inherent characteristics of the entity:
\begin{equation}\label{eq:bench-attr}
    \mathcal{L}, \; \mathcal{S} = \text{MLLM}(\mathcal{P}_\text{attr}, \mathcal{I}),
\end{equation}
where $\mathcal{P}_\text{attr}$ is the prompt for personalized attribute generation, with a concise version:
``\textit{You are generating short- and long-term attributes for a personalized concept based on its original image. Your goal is to define visual traits, behaviors, and preferences that feel natural and realistic.}''
After repeating $N$ times, we construct a personalized concept set $\mathcal{E}_{c} = \{\mathcal{C}_i\}_{i=1}^{N}$ for \B, where each concept $\mathcal{C}$ has \textit{fine-grained} characterizations: $\mathcal{C} = \{\mathcal{I}, \mathcal{S}, \mathcal{L}\}$.

Building upon these personalized concepts with pre-defined both short- and long-term attributes, we now perform \textbf{multi-turn vision-language dialogue contexts construction}.
Specifically, for each personalized concept $\mathcal{C}$, we prompt the MLLM to produce a set of simulated historical vision-language dialogues:
\begin{equation}\label{eq:bench-dialogue}
    \mathcal{D} = \{D_i\}_{i=1}^{{N}_{d}} = \{(Q_{\text{hist},i}, A_{\text{hist},i})\}_{i=1}^{{N}_{d}} = \text{MLLM}(\mathcal{P}_\text{cont}, \mathcal{C}),
\end{equation}
where $D_i = (Q_{\text{hist},i}, A_{\text{hist},i})$ represents each turn of dialogue with a vision-language question $Q_{\text{hist},i}$ and a corresponding answer $A_{\text{hist},i}$, and $\mathcal{P}_\text{cont}$ is the prompt designed for dialogue context generation with a brief version:
``\textit{You are generating multi-turn vision-language dialogue based on the provided concept’s short- and long-term attributions. The dialogue should gradually reveal the concept’s attributes through natural visual conversation.}''
\textit{Notably}, since both short- and long-term attribute variations can lead to changes in the visual appearance of a personalized concept across different dialogue turns, we employ the autoregressive generative model GPT-Image-1~\cite{gptimage1} to generate these images while preserving a consistent visual style.
As illustrated in~\Cref{fig:pipeline-const}, these multi-turn dialogues are designed to include the definition and subsequent modifications of both short- and long-term attributes for that personalized concept, assisting the \A in better understanding and modeling each personalized concept during long-time interactions.
After providing each personalized concept $\mathcal{C}$ with a set of vision-language dialogue contexts $\mathcal{D}$, we obtain a contexts set $\mathcal{E}_{d} = \{\mathcal{D}_i\}_{i=1}^{N} = \{(Q_{\text{hist},i}, A_{\text{hist},i})\}_{i=1}^{N \times N_d}$, where $\mathcal{D}_i$ is the set of historical dialogue contexts designed for personalized concept $\mathcal{C}_i$.

Finally, based on the historical multi-turn vision-language dialogue contexts, we conduct \textbf{evaluation question construction}.
To simulate real-world scenarios where the MLLM can learn from past dialogue contexts to better align with users when referring to their personalized concepts, while also balancing the practicality and feasibility of evaluation, we construct single-turn dialogues (VQA) for evaluation instead of full multi-turn conversations.
Specifically, for each set of multi-turn vision-language dialogue contexts $\mathcal{D} \in \mathcal{E}_d$, we prompt the MLLM to construct several easy and hard VQA questions that require understanding and reasoning over the given context for answering:
\begin{align}\label{eq:bench-eval}
    \mathcal{Q}_{E} &= \{Q_{e,i}, A_{e,i}\}_{i=1}^{{N}_{e}} = \text{MLLM}(\mathcal{P}_\text{easy}, \mathcal{D}),\\
    \mathcal{Q}_{H} &= \{Q_{h,i}, A_{h,i}\}_{i=1}^{{N}_{h}} = \text{MLLM}(\mathcal{P}_\text{hard}, \mathcal{D}),
\end{align}
where $Q_{e,i} = (\mathcal{V}_i, \mathcal{T}_i)$ is an easy VQA question with $\mathcal{V}_i$ denoting the visual input (i.e., an image related to the question) and $\mathcal{T}_i$ denoting the textual input (i.e., the question), and $A_{e,i}$ is the corresponding free-text golden answer. 
Notably, each single-turn dialogue (VQA) $(Q_{e,i}, A_{e,i})$ or $(Q_{h,i}, A_{h,i})$ represents a single complete evaluation. 
The easy and hard VQA questions are defined as follows:
\begin{definition}[Easy VQA Question]
\label{def:easy}
As illustrated in~\Cref{fig:pipeline-const}, an Easy VQA Question targets either short- or long-term attributes of a personalized concept in isolation.  
It requires the MLLM to recognize the concept itself and recall a single type of attribute without jointly reasoning over both time-sensitive and time-stable information.
\end{definition}

\begin{definition}[Hard VQA Question]
\label{def:hard}
As illustrated in~\Cref{fig:pipeline-const}, a Hard VQA Question requires the MLLM to jointly consider and reason over both short- and long-term attributes of a personalized concept.  
Such questions typically involve identifying attribute changes over time, resolving conflicts between persistent and transient preferences, and adapting to contextual shifts based on historical interactions.
\end{definition}
The $\mathcal{P}_\text{easy}$ is the prompt designed for easy VQA question generation with a concise version:
``\textit{You are generating personalized questions based on the dialogue history about a given concept. Each question should be image-grounded and fully answerable based on the historical dialogue, focusing on a single attribute of the concept.}'',
While the $\mathcal{P}_\text{hard}$ prompt builds upon $\mathcal{P}_\text{easy}$ by increasing the difficulty:
\textit{(Same as $\mathcal{P}_\text{easy}$, but) each question should involve multiple short- and long-term attributes and be challenging to answer.}
Combining easy and hard VQA questions for each set of multi-turn vision-language dialogue contexts, we derive the evaluation question set $\mathcal{E}_{e,q} = \{\mathcal{Q}_{E,i}\}_{i=1}^{N} = \{(Q_{e,i}, A_{e,i})\}_{i=1}^{N \times N_e}$ and $\mathcal{E}_{h,q} = \{\mathcal{Q}_{H,i}\}_{i=1}^{N} = \{(Q_{h,i}, A_{h,i})\}_{i=1}^{N \times N_h}$.

\begin{figure}[t]
  \centering
  {\includegraphics[width=0.7\columnwidth]{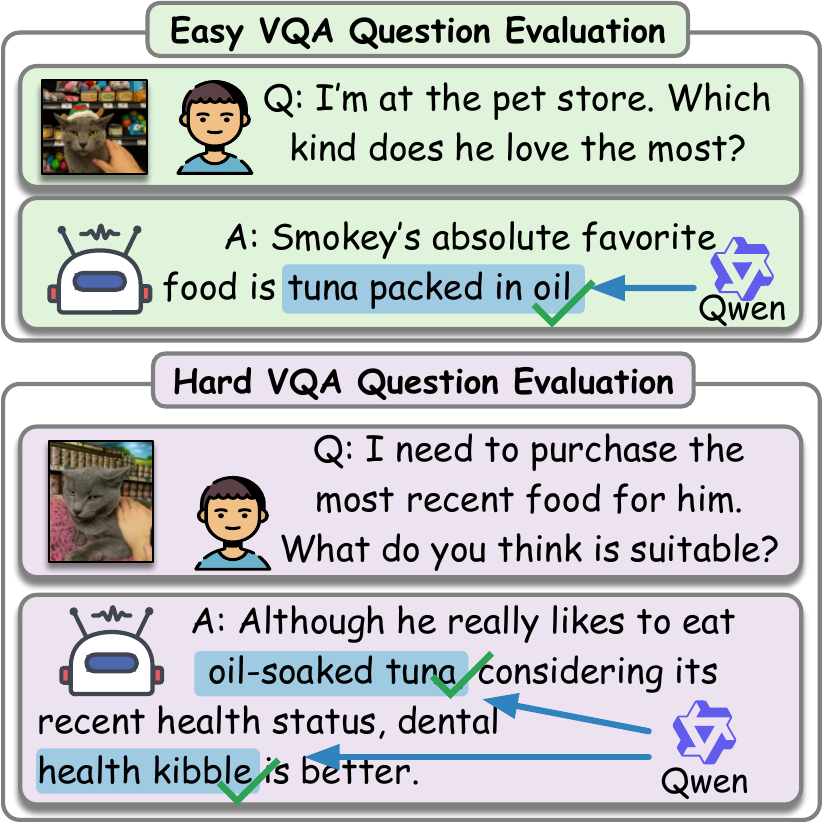}}
  \vspace{-3mm}
  \caption{Evaluation pipeline for \B benchmark.}
  \label{fig:pipeline-eval}
  \vspace{-6mm}
\end{figure}

During \textbf{evaluation}, as illustrated in~\Cref{fig:pipeline-eval}, each assistant is provided with the complete historical vision-language dialogue context and is then required to answer an easy or hard VQA question that is related to a personalized concept.
This setup effectively assesses the capability of \A to learn and interpret variations in each personalized concept from past interactions, and to generate responses that are both personalized and contextually appropriate for the current personalized concept:
\begin{equation}\label{eq:bench-answer}
   \hat{A}_{e/h} =  \operatorname{\Phi_{MLLM}(\mathcal{P}_\text{ans-ex}, \mathcal{E}_d, Q_{e/h}) },
\end{equation}
where $\mathcal{P}_\text{ans-ex}$ is an example prompt designed for generating answers with a brief version: 
``\textit{You are a personalized assistant tasked with handling concept-specific visual questions. Based on the history of personalized dialogue with the user, your role is to identify the concept in question and synthesize both its short- and long-term attributes into a grounded, contextual response.}''.
It is worth noting that, to better simulate challenging and complex real-world applications, we do not explicitly pre-specify which personalized concept is involved in each evaluation question $Q_e$ and only provide the entire past historical context $\mathcal{E}_d$ for the assistant.

As shown in~\Cref{fig:pipeline-eval}, we follow the definition of easy and hard questions (cf. \Cref{def:easy}--\ref{def:hard}) and propose different \textbf{evaluation metrics} for assessing the quality of responses for two kinds of VQA questions.
For the free-text answer $\hat{A}_e$ to an easy question $Q_e$, where only a single type of attribute (either short- or long-term) is involved, we define two metrics: the accuracy (\textbf{ACC-F}) of the free-text response, and the scoring point rate (\textbf{SPR}), which measures the proportion of responses that explicitly reference the relevant attribute variation of the personalized concept.
For the free-text answer $\hat{A}_h$ to a hard question $Q_h$, we retain ACC-F and additionally introduce two metrics: the scoring point rate for correctly referencing long-term attributes (\textbf{SPR-L}) and for short-term attributes (\textbf{SPR-S}), as each question is related to both short- and long-term attributes.
All metrics are automatically evaluated by open-source LLM (Qwen2.5-72B~\cite{yang2024qwen}), where each response is assigned 1 for correctness and 0 otherwise when comparing to the golden answer, and the final score reflects the proportion of correct responses over all evaluated samples.
To enrich the diversity of the evaluation formats, we also provide a single-choice (four-option) version for both easy and hard VQA questions by converting the original free-text answer $A_{e/h}$ into one correct option along with three distractors.
Accordingly, we also define \textbf{ACC-C}, the proportion of correctly selected choices, to quantify the accuracy of this multiple-choice evaluation setting.
We provide additional details, including prompt templates for both construction and evaluation of \B, in~\Cref{app:benchmark_details}.

\begin{figure*}[t]
  \centering
  \includegraphics[width=\linewidth]{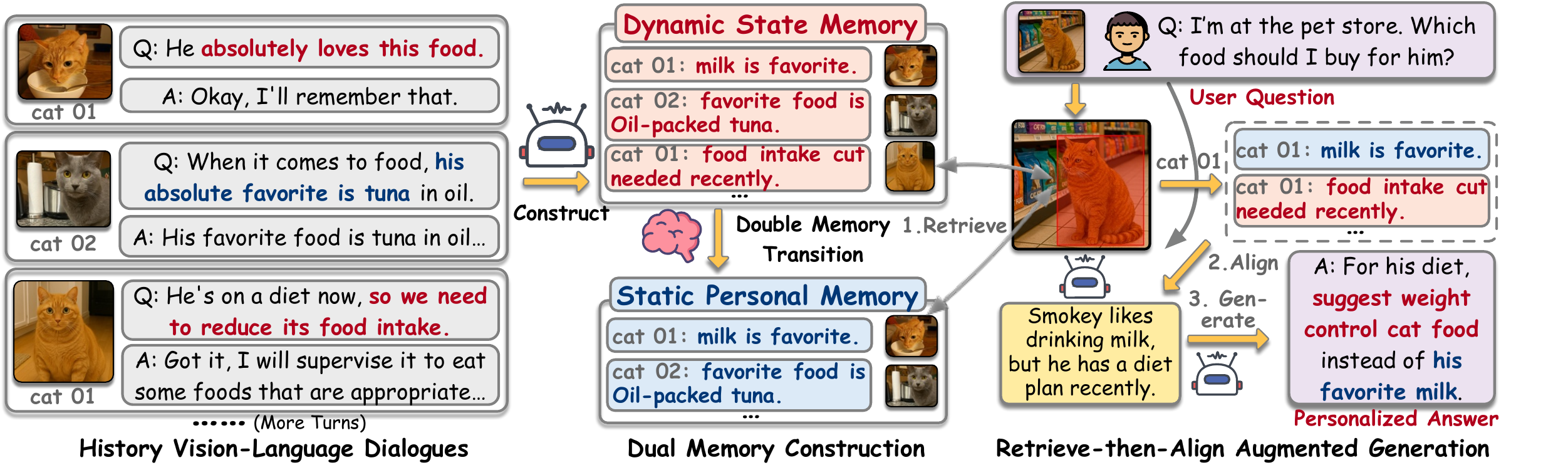}
  \caption{Overall framework of \M, a novel training-free and state-aware personalized MLLM assistant.}
  \label{fig:method}
\end{figure*}

\section{Methodology: \M}
\label{sec:methodology}
\subsection{Overview}
In this section, we present the task definition of \task under our \B, along with the pipeline of our \M.

\noindent \textbf{Task Definition.} 
Unlike prior work that only focuses on context-agnostic simple VQA questions in \task, we provide a new definition under long-context vision-language dialogue scenarios.
The objective of \task under \B is to enable the MLLM to leverage historical multi-turn dialogue contexts involving a user’s personalized concepts, and accurately answer VQA questions grounded in those personalized concepts.
We do not explicitly distinguish the easy and hard VQA questions for simplicity.
We define a set of personalized concepts as $\mathcal{E}_c = \{\mathcal{C}_i\}_{i = 1}^{N}$, and the set of historical multi-turn vision-language dialogue contexts as $\mathcal{E}_d = \{\mathcal{D}_i\}_{i = 1}^{N}$, which encapsulate the visual features (i.e., images) of each personalized concept along with their associated short- and long-term attributes.
Notably, each single-turn history dialogue $(Q_{\text{hist},i}, A_{\text{hist},i})$ also has an $\texttt{Cid}$, which is a concept id uniquely associated with the personalized concept appearing in this dialogue, and will be utilized to identify the dialogue contexts related to this personalized concept. 
A VQA query is defined as $Q = (\mathcal{I}, \mathcal{T}) \in \mathcal{E}_q$, where $\mathcal{I}$ denotes the visual input (i.e., an image relevant to the question), and $\mathcal{T}$ denotes the textual input (i.e., the question itself).
The goal of \A is to generate a contextually fitted personalized free-text answer or an option for $Q$ by considering the whole historical dialogue contexts: $\hat{A} =  \operatorname{\Phi_{MLLM}(\mathcal{P}_\text{ans}, \mathcal{E}_d, Q) }$.

\noindent \textbf{Our Pipeline.}
As illustrated in~\Cref{fig:method}, our \M is the first training-free and state-aware \A.
\M first introduces a \textit{Double-Memory architecture}: a \textit{Dynamic State Memory} that captures and tracks the variations of attributes for personalized concepts (\Cref{subsec:method-dynamic}), and a \textit{Static Personalized Memory} that retains long-term, stable attributes (\Cref{subsec:method-static}).
Besides, a \textit{Double Memory Transition mechanism} is proposed to incrementally transfer long-term attributes from the Dynamic State Memory to the Static Personalized Memory and drop the outdated short-term attributes (\Cref{subsec:method-transition}).
Finally, a training-free Retrieve-then-Align Augmented Generation paradigm is designed to facilitate more contextually fitted personalized generation for queries (\Cref{subsec:method-ra2g}).
The algorithm of our \M is in~\Cref{alg:tame_methodology}.

\subsection{Dynamic State Memory Construction}
\label{subsec:method-dynamic}
To ensure that the assistant can effectively capture the variations of each personalized concept, specifically the changes in their attributes that are critical for improved human experience quality during prolonged interaction, we first introduce a Dynamic State Memory (DS Memory) $\mathcal{M}_{d}$. 
Specifically, for each past vision-language dialogue ($Q_\text{hist} = (\mathcal{I}, \mathcal{T}), A_\text{hist}$), the \A $\Phi_\text{MLLM}$ is instructed to first determine whether the DS Memory requires an update (e.g., an attribute of a personalized concept is newly defined or modified), and then update $\mathcal{M}_{d}$ if necessary:
\begin{equation}\label{eq:method-dynamic}
    \mathcal{M}_{d} \leftarrow \Phi_\text{MLLM}(\mathcal{P}_\text{dsm}, \mathcal{M}_{d}, {Q}_\text{hist}, A_\text{hist}),
\end{equation}
where $\mathcal{P}_\text{dsm}$ is the prompt designed for the update of DS Memory through three atomic operations~\cite{zhao2024expel}: \textbf{Add}: Add a new attribute for a personalized concept, \textbf{Modify}: Modify an existing attribute, and \textbf{Remove}: Remove an existing attribute, with a concise version:
``\textit{You are maintaining a dynamic state memory for a specific concept. Your task is to analyze historical dialogues to identify and extract visual traits or behavioral attributes of the concept. Add new attributes when newly revealed, modify existing ones if conflicts arise, and remove when necessary.}''
Each memory item $m_{d} \in \mathcal{M}_{d}$ is stored in the format:
$m_{d} = \;<\texttt{Cid},\ \texttt{Attr}>,$
where $\texttt{Cid}$ is a concept id uniquely associated with each personalized concept, and $\texttt{Attr}$ is the attribute record which may contain both textual and visual information of the attribute associated with that $\texttt{Cid}$.

\subsection{Static Personalized Memory Construction}
\label{subsec:method-static}
To improve the efficiency of the DS Memory and better manage the two types of attributes, we further introduce a Static Personalized Memory (SP Memory) $\mathcal{M}_{s}$.
In contrast to the DS Memory, the SP Memory is not directly updated based on historical dialogues. Instead, it is modified only through transitions from the DS Memory, serving to stabilize long-term attribute representations:
\begin{equation}\label{eq:method-static}
    \mathcal{M}_{s} \leftarrow \Phi_\text{MLLM}(\mathcal{P}_\text{tran}, \mathcal{M}_{d}, \mathcal{M}_{s}), \quad \text{if } \psi_\text{trigger}(\mathcal{M}_{d}) = \text{True},
\end{equation}
where $\mathcal{P}_\text{tran}$ is the prompt designed for transferring the memory item from the DS to the SP Memory using the same three atomic operations defined in the prior section, with a brief version: 
``\textit{You’re managing a transition from dynamic state memory to static personalized memory.
Your task is to ensure efficiency by identifying persistent, stable information within dynamic memory, removing it, and appending it to static personalized memory.}''
And $\psi_\text{trigger}$ is designed for the double memory transition mechanism that will be discussed in the next section.
Each memory item $m_{s} \in \mathcal{M}_{s}$ has a similar structure as the DS Memory:
$m_{s} = \;<\texttt{Cid},\; \texttt{Attr}>$,
where each $\texttt{Attr}$ is a long-term attribute for the personalized concept with that $\texttt{Cid}$.

\subsection{Double Memory Transition}
\label{subsec:method-transition}
In this section, we detail the Double Memory Transition mechanism.
Specifically, we define the trigger function $\psi_\text{trigger}(\cdot)$ as follows:
\begin{equation}\label{eq:method-transition}
    \psi_\text{trigger}(\mathcal{M}_d) \triangleq 
    \begin{cases}
        \text{True}, & \text{if } |\mathcal{M}_d| > \tau; \\
        \text{True}, & \text{if } \exists \, m_d \in \mathcal{M}_d, \text{Long-Term}(m_d) = \text{True}; \\
        \text{False}, & \text{otherwise}.
    \end{cases}
\end{equation}
Here, $\tau$ is the maximum size of the DS Memory, which is set to ensure both efficiency and temporal relevance.
After each update of the DS Memory with historical dialogue contexts, the trigger function is first invoked to check whether any memory item $m_d$ contains a long-term attribute. 
If this condition is satisfied, a transition operation is triggered (cf.~\cref{eq:method-static}) to transfer the relevant item into the SP Memory.
Subsequently, the trigger function is executed again by \M to examine whether the memory size exceeds $\tau$.
If this is true, a First In, First Out (FIFO) policy is applied to discard the outdated short-term attributes for that personalized concept.

\begin{algorithm}[t]
\caption{\M Algorithm}
\label{alg:tame_methodology}
\begin{algorithmic}[1]
\Require Historical dialogue contexts $\mathcal{E}_d$, User query $Q = (\mathcal{I}, \mathcal{T})$, MLLM $\Phi_\text{MLLM}$, Memory size threshold $\tau$, Similarity threshold $\phi$, Category set $\mathcal{K}$
\Ensure Personalized answer $\hat{A}$, Updated memories $\mathcal{M}_d$ and $\mathcal{M}_s$

\State \textbf{Initialize:} $\mathcal{M}_d \gets \emptyset$, $\mathcal{M}_s \gets \emptyset$

\State \textbf{/* Phase 1: Double Memory Construction*/}
\For{each historical dialogue $(Q_\text{hist}, A_\text{hist}) \in \mathcal{E}_d$}
    \State $\mathcal{M}_d \leftarrow \Phi_\text{MLLM}(\mathcal{P}_\text{dsm}, \mathcal{M}_d, Q_\text{hist}, A_\text{hist})$
    
    \If{$\psi_\text{trigger}(\mathcal{M}_d) = \text{True}$}
        \If{$\exists \, m_d \in \mathcal{M}_d, \text{Long-Term}(m_d) = \text{True}$}
        \State $\mathcal{M}_s \leftarrow \Phi_\text{MLLM}(\mathcal{P}_\text{tran}, \mathcal{M}_d, \mathcal{M}_s)$
        \EndIf
        \If{$|\mathcal{M}_d| > \tau$}
         \State Remove outdated short-term attributes
        \EndIf
    \EndIf
\EndFor

\State \textbf{/*Phase 2: Retrieve-then-Align Augmented Generation*/}
\State $\{\mathcal{I}_{\text{ent}_1}, \mathcal{I}_{\text{ent}_2}, \ldots, \mathcal{I}_{\text{ent}_n}\} \gets \text{Segment}(\mathcal{I}, \mathcal{K})$

\For{each entity image $\mathcal{I}_{\text{ent}}$ in segmented images}
    \State $\texttt{Cid} \gets \operatorname*{arg\,max}_{(\mathcal{I}_k, \mathcal{T}_k) \in \mathcal{M}_d \cup \mathcal{M}_s}$
    \State \hspace{1cm} $(\frac{\Phi_v(\mathcal{I}_{\text{ent}}) \cdot \Phi_v(\mathcal{I}_k)}{\|\Phi_v(\mathcal{I}_{\text{ent}})\| \, \|\Phi_v(\mathcal{I}_k)\|} + \frac{\Phi_t(\mathcal{T}) \cdot \Phi_t(\mathcal{T}_k)}{\|\Phi_t(\mathcal{T})\| \, \|\Phi_t(\mathcal{T}_k)\|})$

    \State Locate memory $\mathcal{M}_d^{\texttt{c}} \gets \{\, (\texttt{Cid},\texttt{Attr}) \in \mathcal{M}_d \mid \texttt{Cid} = \texttt{c} \,\}$
    \State Locate memory $\mathcal{M}_s^{\texttt{c}} \gets \{\, (\texttt{Cid},\texttt{Attr}) \in \mathcal{M}_s \mid \texttt{Cid} = \texttt{c}\,\}$
    \State Collect these two memories to form $\mathcal{M}_{\text{f}}^{\texttt{c}}$
    
    \State $\mathcal{H} \gets \Phi_\text{MLLM}(\mathcal{P}_\text{align}, Q, \mathcal{M}_{\text{f}})$
\EndFor

\State $\hat{A} \gets \Phi_\text{MLLM}(\mathcal{P}_\text{ans}, Q, \mathcal{H})$

\State \textbf{/* Phase 3: Memory Update with Current Interaction */}
\State Add $(Q, \hat{A})$ to historical dialogues
\State Update $\mathcal{M}_d$ using \cref{eq:method-dynamic}: 
\State \hspace{1cm} $\mathcal{M}_d \leftarrow \Phi_\text{MLLM}(\mathcal{P}_\text{dsm}, \mathcal{M}_d, Q, \hat{A})$

\State \Return $\hat{A}$, $\mathcal{M}_d$, $\mathcal{M}_s$
\end{algorithmic}
\end{algorithm}

\subsection{Retrieve-then-Align Augmented Generation}
\label{subsec:method-ra2g}
Mainstream works in \task typically fine-tune MLLMs for personalized generation~\cite{alaluf2024myvlm, hao2024rap, pi2025personalized}, which is costly and infeasible for closed-source MLLMs.
To overcome this limitation and better adapt our double-memory framework, inspired by the RAG technique in modern LLMs, we propose a novel training-free Retrieve-then-Align Augmented Generation (\RAG) paradigm.
This paradigm first retrieves relevant personalized concept information from multiple memories, then aligns it with the user query to extract the most contextualized knowledge for generation.
Specifically, for each user question $Q = (\mathcal{I}, \mathcal{T})$, we first employ the zero-shot open-set grounding model to localize entities and segment them in the original query image $\mathcal{I}$, generating refined entity-wise images:
\begin{equation}\label{eq:segment}
   \text{Segment}(\mathcal{I}, \mathcal{K}) = \{ \mathcal{I}_{\text{ent}_1}, \mathcal{I}_{\text{ent}_2}, \ldots, \mathcal{I}_{\text{ent}_n} \},
\end{equation}
where $\mathcal{K}$ is a predefined category set designed to cover nearly all entity types.
To simplify the narrative, we consider the case that only one entity image $\mathcal{I}_{\text{ent}}$ is derived (i.e., only one personalized concept is related), but \M can handle the multi-concept scenarios.
We then compare $\mathcal{I}_\text{ent}$ with the visual information $\mathcal{I}_k$ (i.e., images) and textual information $\mathcal{T}_k$ (i.e., attributes) from each item $m_k$ in both memories to identify the corresponding \texttt{Cid} for the personalized concept in $\mathcal{I}_{\text{ent}}$:
\begin{equation}\label{eq:method-retrieve}
    m_{*}  =  \operatorname*{arg\,max}_{m_k \in (\mathcal{M}_d \cup \mathcal{M}_s)} 
    (\frac{\Phi_v(\mathcal{I}_{\text{ent}}) \cdot \Phi_v(\mathcal{I}_k)}
    {\|\Phi_v(\mathcal{I}_{\text{ent}})\| \, \|\Phi_v(\mathcal{I}_k)\|} \\
     + \frac{\Phi_v(\mathcal{I}_{\text{ent}}) \cdot \Phi_t(\mathcal{T}_k)}
    {\|\Phi_v(\mathcal{I}_{\text{ent}})\| \, \|\Phi_t(\mathcal{T}_k)\|}),
\end{equation}
where $\texttt{c}$ is the retrieved $\texttt{Cid}$ in $m_{*}$ for $\mathcal{I}_{\text{ent}}$, while $\Phi_v$ and $\Phi_t$ are embedding models for retrieval.
We then locate memory items $\mathcal{M}^c = \mathcal{M}_d^{\texttt{c}} \cup \mathcal{M}_s^{\texttt{c}}$ from $\mathcal{M}_d \cup \mathcal{M}_s$ that share the same \texttt{Cid = c}, and retrieve the most relevant $E$ items with respect to the user query from the located memories $\mathcal{M}^c$ if the number of items in $\mathcal{M}^c$ exceeds $E$, forming a set of valuable contexts $\mathcal{M}_f^{\texttt{c}}$.
The contexts $\mathcal{M}_f^{\texttt{c}}$ are then fed into the MLLM with the user question for alignment, which aims to extract the most relevant information while filtering out noise from the multi-memories:
\begin{equation}\label{eq:method-align}
    \mathcal{H} = \Phi_\text{MLLM}(\mathcal{P}_\text{align}, Q, \mathcal{M}_f^{\texttt{c}}),
\end{equation}
where $\mathcal{H}$ denotes the user-query aligned textual contextual information derived from historical vision-language dialogues, and $\mathcal{P}_\text{align}$ is the prompt designed for alignment with a concise version: 
``\textit{You are extracting information aligned with the user’s question from concept-related historical context to support accurate and relevant answers.}''.
\begin{table*}[t]
    \caption{Performance comparison on the \B-E and \B-H datasets. The best results are in \textbf{black bold}, while the second-best are \underline{underlined}. 
    Higher values of ACC-C, ACC-F, SPR, SPR-L, and SPR-S indicate better performance.}
    \label{tab:main-result}
    \vspace{-2mm}
        \renewcommand{\arraystretch}{1.1}
    \centering
    \setlength{\tabcolsep}{6pt}
    \resizebox{\linewidth}{!}{
    \begin{tabular}{l|c|c|c|ccc|cccc}
    \shline
     \multirow{2}{*}{\textbf{Methods}} & \multirow{2}{*}{\textbf{MLLM Backbone}} & \multirow{2}{*}{\makecell[c]{\textbf{Training}  \textbf{Free}}} & \multirow{2}{*}{\makecell[c]{\textbf{State}  \textbf{Aware}}}  & \multicolumn{3}{c|}{\textbf{\B-E}} & \multicolumn{4}{c}{\textbf{\B-H}} \\ 
     & & & & \textbf{ACC-C} &  \textbf{ACC-F} &  \textbf{SPR} &  \textbf{ACC-C} &  \textbf{ACC-F} &  \textbf{SPR-L} &  \textbf{SPR-S} \\
    \hline 
    \multirow{2}{*}{\makecell[c]{MyVLM}} &  LLaVA-v1.6-vicuna-7B & \multirow{2}{*}{\makecell[c]{\ding{55}}} & \multirow{2}{*}{\makecell[c]{\ding{55}}} & 39.16 & 9.58 & 14.58 & 32.22 & 8.33 & 11.11 & 12.78 \\
    &  LLaVA-v1.6-mistral-7B &  & & 42.91 & 10.83 & 13.75 & 35.56 & 15.56 & 17.22 & 18.89 \\ \hdashline 
    \multirow{2}{*}{\makecell[c]{Yo'LLaVA}} & LLaVA-v1.5-7B & \multirow{2}{*}{\makecell[c]{\ding{55}}} & \multirow{2}{*}{\makecell[c]{\ding{55}}} & 40.41 & 6.47 & 16.81 & 28.89 & 10.92 & 10.34 & 19.54 \\
    &  LLaVA-v1.5-13B &  & & 42.08 & 5.42 & 15.83 & 30.56 & 14.44 & 11.67 & 21.11 \\ \hdashline 
    \multirow{2}{*}{\makecell[c]{RAP}} &  LLaVA-v1.5-13B & \multirow{2}{*}{\makecell[c]{\ding{55}}} & \multirow{2}{*}{\makecell[c]{\ding{55}}} & 46.66 & 4.18 & 31.38 & 39.11 & 3.91 & 36.31 & 17.88 \\
    &  Phi3-V-3.8B &  & & 52.08 & 12.13 & 19.25 & 45.81 & 7.82 & 10.61 & 7.26 \\ \hdashline 

    \multirow{2}{*}{\makecell[c]{PeKit}} & Qwen2.5-VL-7B & \multirow{2}{*}{\makecell[c]{\ding{51}}} & \multirow{2}{*}{\makecell[c]{\ding{55}}} & 53.33 & 25.83 & 30.83 & 46.11 & 17.78 & 16.11 & 20.00 \\
    &  InternVL3-8B &  & & 56.25 & 27.92 & 28.33 & 42.22 & 22.22 & 17.78 & 30.56 \\ \hdashline  

    \multirow{2}{*}{\makecell[c]{R2P}} & Qwen2.5-VL-7B & \multirow{2}{*}{\makecell[c]{\ding{51}}} & \multirow{2}{*}{\makecell[c]{\ding{55}}} & 44.17 & 38.33 & 44.58 & 39.44 & 33.33 & 40.55 & 37.22 \\
    &  InternVL3-8B &  & & 43.75 & 40.41 & 42.91 & 39.44 & 38.33 & 42.22 & 40.55 \\ \hline

    \multirow{2}{*}{\makecell[c]{\textbf{\M}}} & Qwen2.5-VL-7B & \multirow{2}{*}{\makecell[c]{\ding{51}}} & \multirow{2}{*}{\makecell[c]{\ding{51}}}    & \underline{75.41} & \underline{46.11} & \underline{47.08} & \underline{66.11} & \underline{46.11} & \underline{43.88} & \underline{49.44}\\ 
    & InternVL3-8B & & & \textbf{77.08} & \textbf{52.22} & \textbf{58.75} & \textbf{69.44} & \textbf{56.66} & \textbf{53.33} & \textbf{56.66 }\\ 
    
    \shline
    \end{tabular}
    }
\end{table*}
Finally, the aligned contextual information is fed with the user query to generate the contextually fitted response:
\begin{equation}\label{eq:method-answer}
    \hat{A} = \Phi_\text{MLLM}(\mathcal{P}_\text{ans}, Q, \mathcal{H}).
\end{equation}
After answering user queries, the resulting query-answer pairs become part of the historical dialogue and is used by \M to update the double memories.

\section{Experiments}
\label{sec:experiments}
\subsection{Experimental Setup}
In this section, we present a concise version of the experimental setup, with a detailed version at~\Cref{app:exp_setup}.

\noindent \textbf{Baselines.}
To assess the effectiveness of \M, we compare it with several competitive baselines in \task, including MyVLM~\cite{alaluf2024myvlm}, Yo'LLaVA~\cite{nguyen2024yollava}, RAP~\cite{hao2024rap}, PeKit~\cite{seifi2025personalization}, and R2P~\cite{das2025training}.

\noindent \textbf{Benchmarks and Metrics.}
We evaluate all models on our proposed \textbf{\B} benchmark, which consists of 30 personalized concepts, including Pet (10), Object (10), and Person (10). 
We generate approximately 420 historical dialogues, 240 simple questions, 180 hard questions, and 720 personalized images.
As mentioned in~\Cref{sec:benchmark}, \B has two subsets: \textbf{\B-E}, containing \underline{E}asy VQA questions involving single-type attributes, and \textbf{\B-H}, containing \underline{H}ard questions that involve reasoning over double-type attributes.
For \B-E, we introduce three evaluation metrics: Accuracy of Choice questions (\textbf{ACC-C}), Accuracy of Free-text questions (\textbf{ACC-F}), and Scoring Point Rate (\textbf{SPR}).
For \B-H, we refine SPR into two metrics, SPR-Long-term Attribute (\textbf{SPR-L}) and SPR-Short-term Attribute (\textbf{SPR-S}), to better accommodate questions that require reasoning over both types of attributes.
Notably, while \B focuses on the long-context aspect of \task, its VQA questions also involve assessing the capabilities of \A emphasized in prior benchmarks (e.g., visual identification and textual replacement of personalized concepts).

\begin{figure*}[t]
  \centering
  \includegraphics[width=0.97\linewidth]{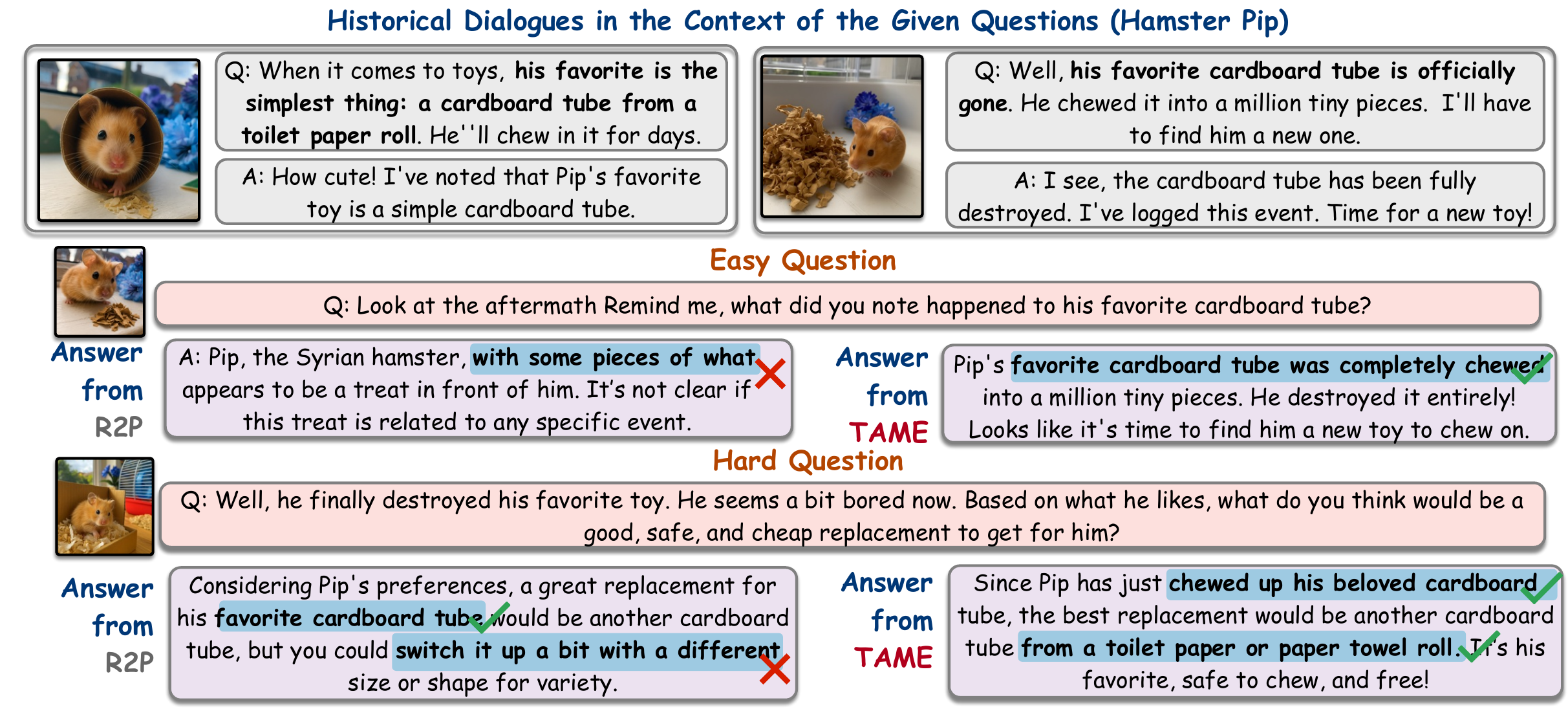}
    \vspace{-2mm}
  \caption{Case study of easy and hard VQA in \B between \M and competitive baseline R2P.}
  \label{fig:case-study}
\end{figure*}

\noindent \textbf{Implementation Details.}
In this study, we employ the MLLM \texttt{Qwen2.5-VL-7B}~\cite{bai2025qwenvl} and \texttt{InternVL3-8B}~\cite{zhu2025internvl} as backbone for \M.  
The pre-defined maximum size $\tau$ for the DS Memory is set to 10.  
To retrieve related concepts, we use the \texttt{Grounding-DINO}~\cite{liu2024grounding} model for open-set grounding on the query image, and adopt the \texttt{Jina-Embedding-v4}~\cite{günther2025jinaembeddingsv} to generate embeddings for both visual and textual inputs.  
We define a fixed category set $\mathcal{K} = \{\texttt{person}, \texttt{animal}, \texttt{household item}, \texttt{personal belonging}\}$ to filter candidates.
The details for implementing the baselines on \B are in~\Cref{app:exp_setup}.

\subsection{Main Performance}
We compare \M with 5 competitive baselines on \B, including \B-E and \B-H datasets, with results reported in~\Cref{tab:main-result}.
From the results, we have the following observations.
Repetitive training-based methods such as MyVLM and Yo’LLaVA perform worse on both datasets, as storing abundant contextual information in a purely parametric manner proves ineffective when facing complex, information-rich scenarios.
For baseline RAP, which only requires a single round of training and incorporates memory mechanisms to store personalized information for subsequent training-free inference, demonstrates moderate improvements.
However, their dataset-specific pre-training limits generalization to newly emerging personalized concepts and scenarios, particularly in cases where user queries are state-aware and demand long-context understanding rather than simple visual identification.
Baselines PeKit and R2P, which employ the RAG mechanism for training-free \task, achieve promising performance on both datasets. 
This is largely attributed to the RAG mechanism to store and leverage abundant historical dialogue contexts for state-aware personalized generation.
However, storing the variations of both short- and long-term attributes of personalized concepts in a single memory complicates the retrieval process when searching for the most appropriate context for generation, especially in cases where both types of attributes are required.
These drawbacks significantly limit their performance on \B-H.

In contrast, our \M achieves the best performance on both datasets, providing the highest experience quality under the long-context \task scenarios.
These performance improvements stem from the superior design of the double-memory mechanism, which equips \A with more fine-grained memory management capabilities to handle the complex real-world variations within each personalized concept.
Moreover, the \RAG paradigm enables more question-aligned contexts to be derived from multi-memory knowledge, facilitating more appropriate and well-adapted personalized generation.
As illustrated in~\Cref{fig:case-study}, compared to R2P baseline, \M provides more accurate and fitted personalization for complicated user queries that contain complex variations of personalized concept attributes. 

\begin{table}[t]
    \centering
    \caption{Ablation study on core components in \M.}
    \vspace*{-2mm}
    \label{tab:ablation}
        \setlength{\tabcolsep}{4.5pt}
    \resizebox{\linewidth}{!}{
    \begin{tabular}{cccccc}
        \toprule
         & & \multicolumn{4}{c}{\textbf{LCMP-H}}\\
       \textbf{Module} &  \textbf{Variant} & \textbf{ACC-C} & \textbf{ACC-F} &\textbf{SPR-L} & \textbf{SPR-S} \\
        \midrule
     \multirow{3}{*}{\makecell[c]{Double \\ Memory}} 
    & w/o DS Memory  & 64.44 & 43.33 & 40.00 & 47.78  \\ 
    & w/o SP Memory & 66.11 & 43.89 & 43.33 & 54.44  \\
    & w/o Memory & 65.55 & 42.78 & 40.00 & 49.44  \\
    \midrule
    \multirow{2}{*}{\makecell[c]{\RAG}} 
    & w/o Alignment & 63.88 & 41.11 & 47.78 & 48.33   \\
    & w/o Retrieval & 67.22 & 48.33 & 51.66 & 49.44  \\
    \midrule
    \rowcolor{gray!12}
    \textbf{\M} & \textbf{ALL}  & \textbf{69.44} & \textbf{56.66} & \textbf{53.33} & \textbf{56.66}  \\ 
     \bottomrule
    \end{tabular}
    }
\end{table}

\subsection{Ablation Study}
We conduct an ablation study to assess the role of each component in \M on \B-H dataset, and the results are presented in~\Cref{tab:ablation}.
To evaluate the \textit{Double-Memory mechanism}, we design three variants:
\textbf{(1) w/o DS Memory}, which eliminates the DS Memory, \textbf{(2) w/o SP Memory}, which removes the SP Memory, and \textbf{(3) w/o Memory}, where both memories are removed.
We observe that eliminating either type of memory leads to a significant performance drop in \M, as the absence of either short- or long-term attribute storage results in incomplete context during generation.
Moreover, removing the memory mechanism leads to a significant performance degradation, as directly retrieving from abundant dialogue contexts without a well-structured organization becomes increasingly complex and noisy.
To validate the \textit{\RAG paradigm}, we introduce two variants: \textbf{(1) w/o Alignment}, where the retrieved multi-memory contextual information is directly leveraged for personalized generation without the alignment process, and \textbf{(2) w/o Retrieval}, where the abundant context information of all personalized concepts is first abbreviated and then fed into \M one-time for alignment and generation without any filtering.
We observe a clear performance decline in the variant w/o Alignment, as simply providing all retrieved contexts from multi-memory to the MLLM assistant introduces substantial noise under complex real-world user queries (i.e., both types of personalized concept attributes varying in \B-H dataset). 
These irrelevant contexts can mislead the MLLM and result in inaccurate responses.
Moreover, removing the retrieval module also leads to a noticeable performance decline, as incorporating excessive irrelevant or even noisy contexts complicates both the alignment and generation processes.

\section{Conclusion}
\label{sec:conclusion}
In this work, we introduced the first benchmark (\B) to quantitatively evaluate the interaction quality of \A under long-context scenarios, focusing on the capability of MLLMs in capturing the complex variations in personalized concepts in multi-turn vision-language dialogues for more contextually fitted responses.
Besides, we proposed \M, a novel baseline for \B that is training-free and state-aware.
\M equips \A with double memories to effectively manage variations in personalized concepts across different time scales.
Moreover, the Retrieve-then-Align Augmented Generation paradigm matches the most contextually relevant information from multi-memory retrieval, facilitating contextually appropriate personalized generation.
Experiments and cases on \B show that \M achieves superior performance, highlighting its potential as a promising state-aware MLLM assistant for future applications.

\section*{Acknowledgments}
This work was supported by National Natural Science Foundation of China (Grant No. 62572097, No. 62176043, and No. U22A2097).

\newpage

\bibliographystyle{ACM-Reference-Format}
\balance
\bibliography{main}

\newpage

\clearpage
\setcounter{page}{1}
\renewcommand{\thefigure}{S\arabic{figure}}
\setcounter{figure}{0}
\appendix

\section{More Details about \B}
\label{app:benchmark_details}

\subsection{Algorithm of Construction}
\label{app:alg_dataset}
As shown in \Cref{alg:benchmark_construction}, we provide a complete algorithm procedure for constructing the \B benchmark.

\subsection{Presentation of Construction Prompts}
In this section, we provide complete prompts employed in personalized concept construction (\Cref{prompt:concept}), multi-turn vision-language contexts construction (\Cref{prompt:dialogue}), evaluation question construction (\Cref{prompt:easy,prompt:hard,prompt:hard-2}), evaluating free-text answers (\Cref{prompt:eval-freetext}), and evaluating scoring points (\Cref{prompt:eval-scoring}). 
Further, we also provide a prompt template (\Cref{prompt:image-gen}) to generate contextual personalized images.

\subsection{Construction Details}
To enrich diversity and personalization in personalized concepts generation, we supplement the Microsoft COCO dataset with a small set of manually collected raw images focusing on pet and object concepts. 

We then employ a set of advanced pre-trained generative models to construct personalized concepts with their associated dialogues and evaluations.
Specifically, we utilize \texttt{Gemini-2.5-Pro}~\cite{gemini2.5pro}, a state-of-the-art reasoning model with visual understanding, to simulate realistic personalized scenarios. 
For consistent and natural personalized image synthesis, we leverage \texttt{GPT-Image-1}~\cite{gptimage1}, an auto-regressive image generation model, which conditions on the raw concept images together with contextual queries. 
These powerful closed-source models are accessed via their official APIs.

For free-text and scoring point evaluations, we employ an open-source LMM (\texttt{Qwen-2.5-72B-Instruct}~\cite{yang2024qwen}) to provide robust and reproducible evaluation results. 
The generations are configured with a temperature setting of 0.8.

\subsection{Concepts Overview in \B}
\begin{figure*}[p]
  \centering
  \includegraphics[width=0.70\linewidth]{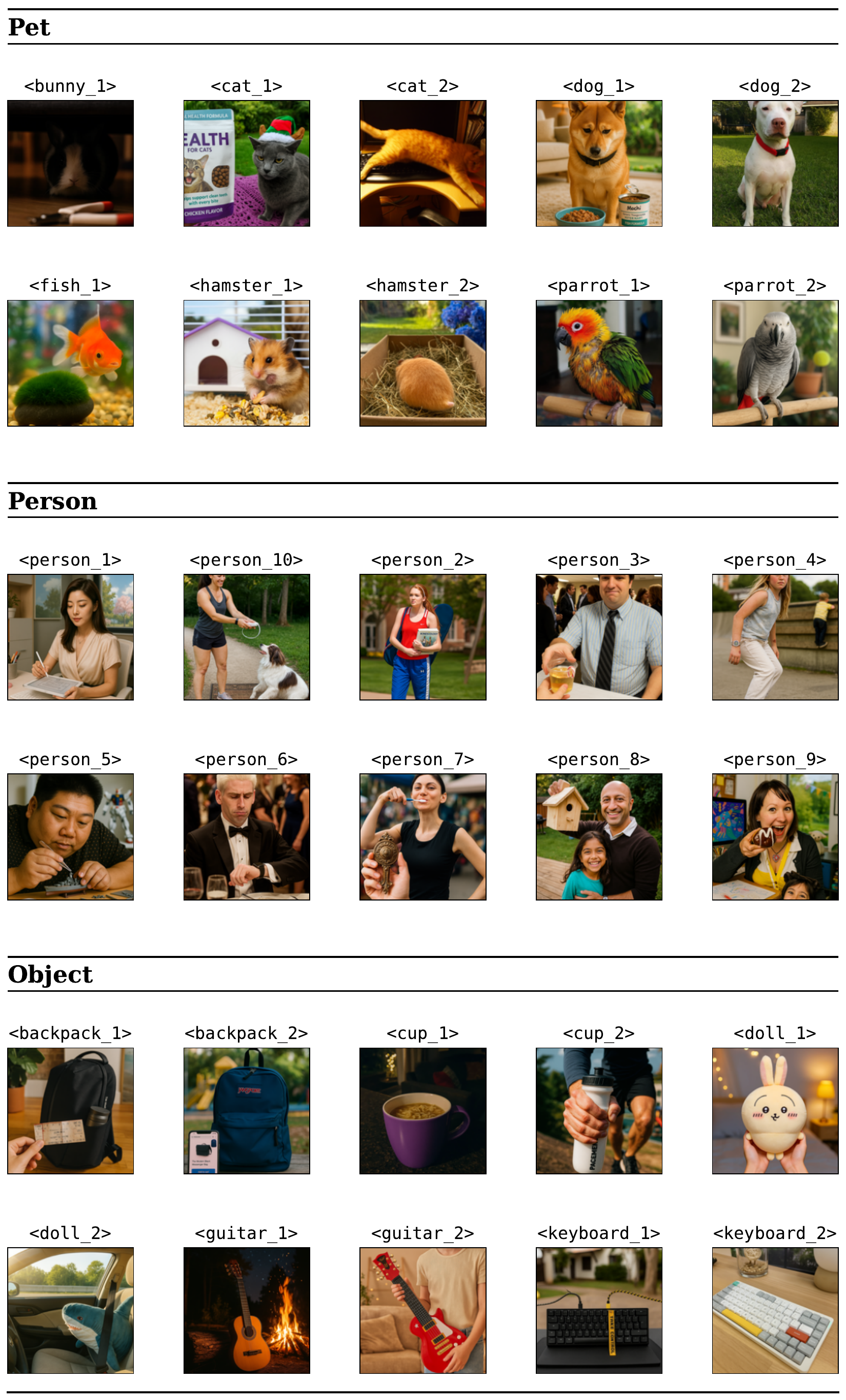}
  \vspace{-3mm}
  \caption{Personalized Concepts Overview.}
  \label{fig:app-concept-overview}
  \vspace{-3.5mm}
\end{figure*}

In this section, we present the portrait of all concepts in \Cref{fig:app-concept-overview}, including Pet (10), Object (10), and Person (10), constructed in the proposed \B benchmark.

\section{Case Study of Concepts in \B}
\label{app:benchmark-case}
\begin{figure*}[p]
  \centering
  \includegraphics[width=0.70\linewidth]{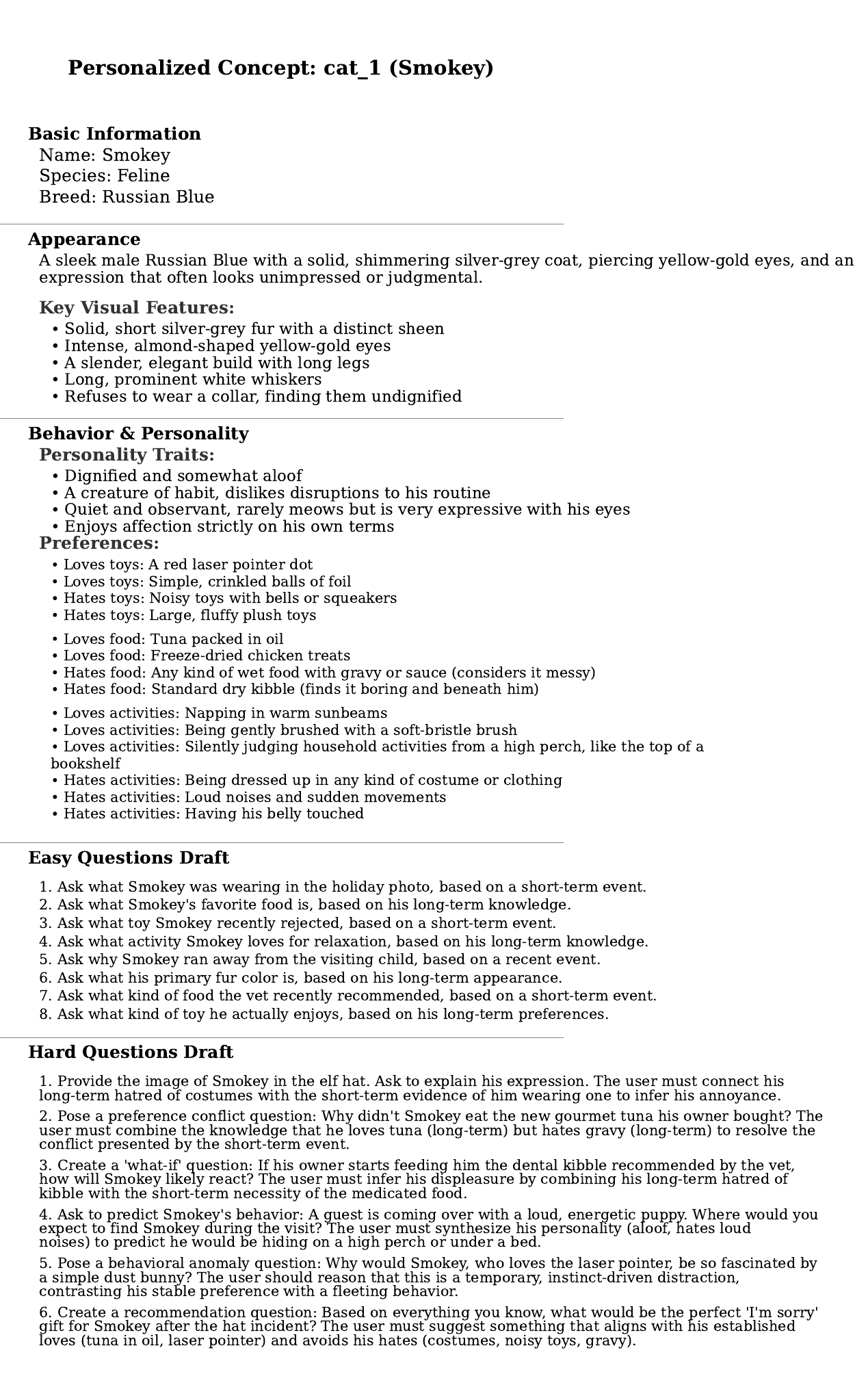}
  \vspace{-4mm}
  \caption{Presentation of Concept content for Personalized Concept (Cat 1).}
  \label{fig:app-concept-attributes-cat1}
\end{figure*}

\begin{figure*}[p]
  \centering
  \includegraphics[width=0.90\linewidth]{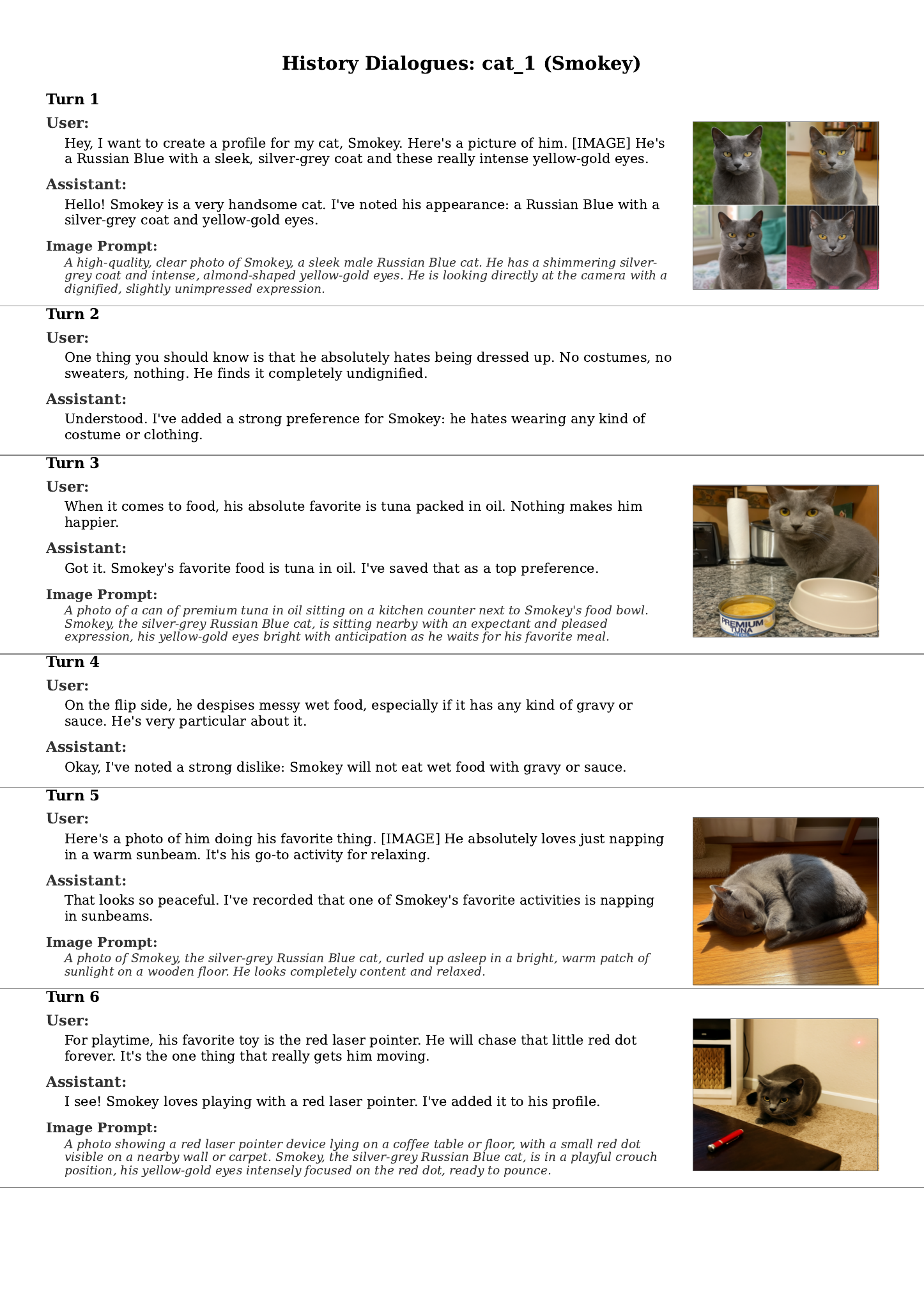}
  \vspace{-20mm}
  \caption{Presentation of Dialogue Contexts for Personalized Concept (Cat 1).}
  \label{fig:app-concept-history-cat1}
\end{figure*}

\begin{figure*}[p]
  \centering
  \includegraphics[width=0.90\linewidth]{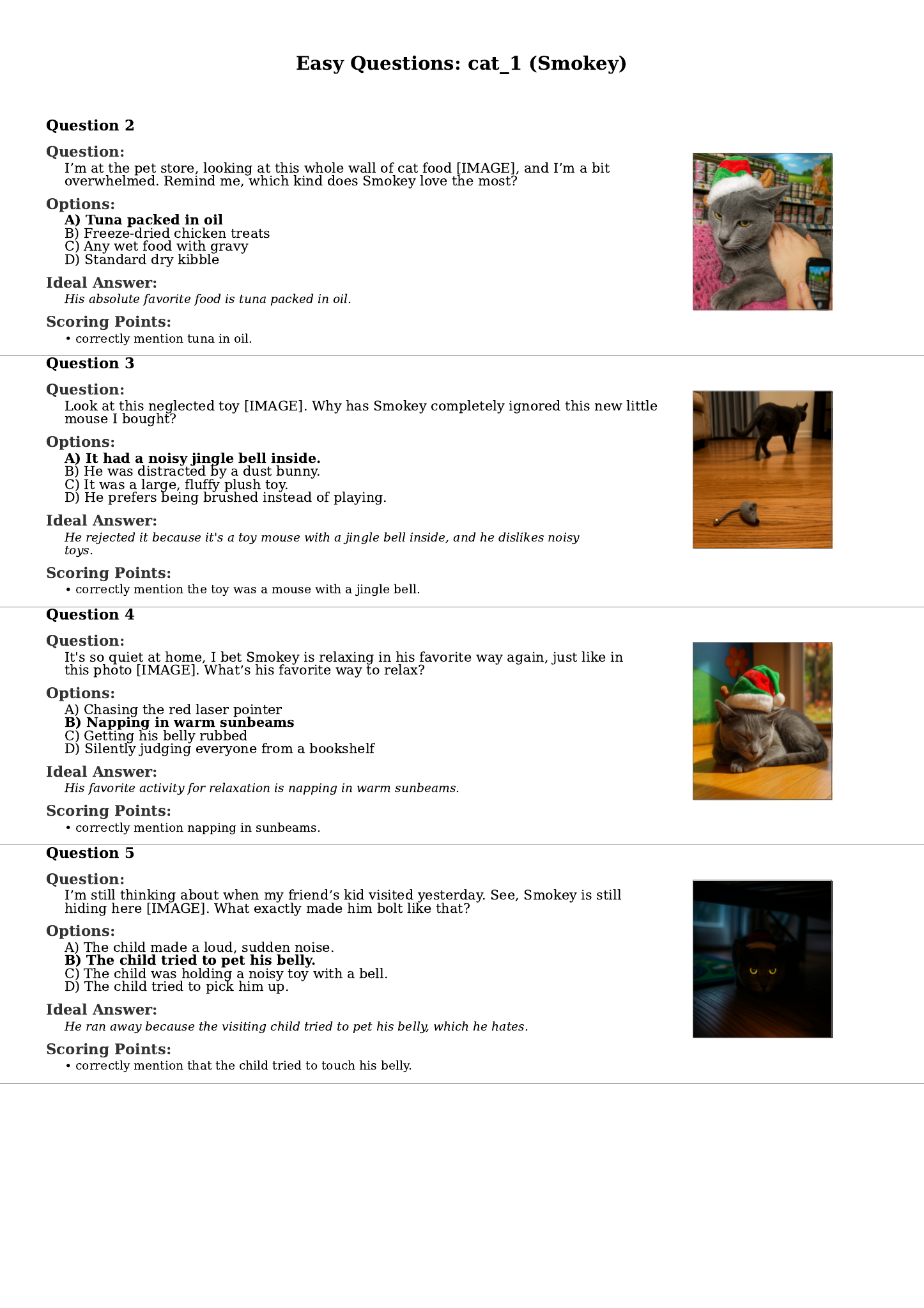}
  \vspace{-30mm}
  \caption{Presentation of Easy Questions for Personalized Concept (Cat 1).}
  \label{fig:app-concept-easy-cat1}
\end{figure*}

\begin{figure*}[p]
  \centering
  \includegraphics[width=0.90\linewidth]{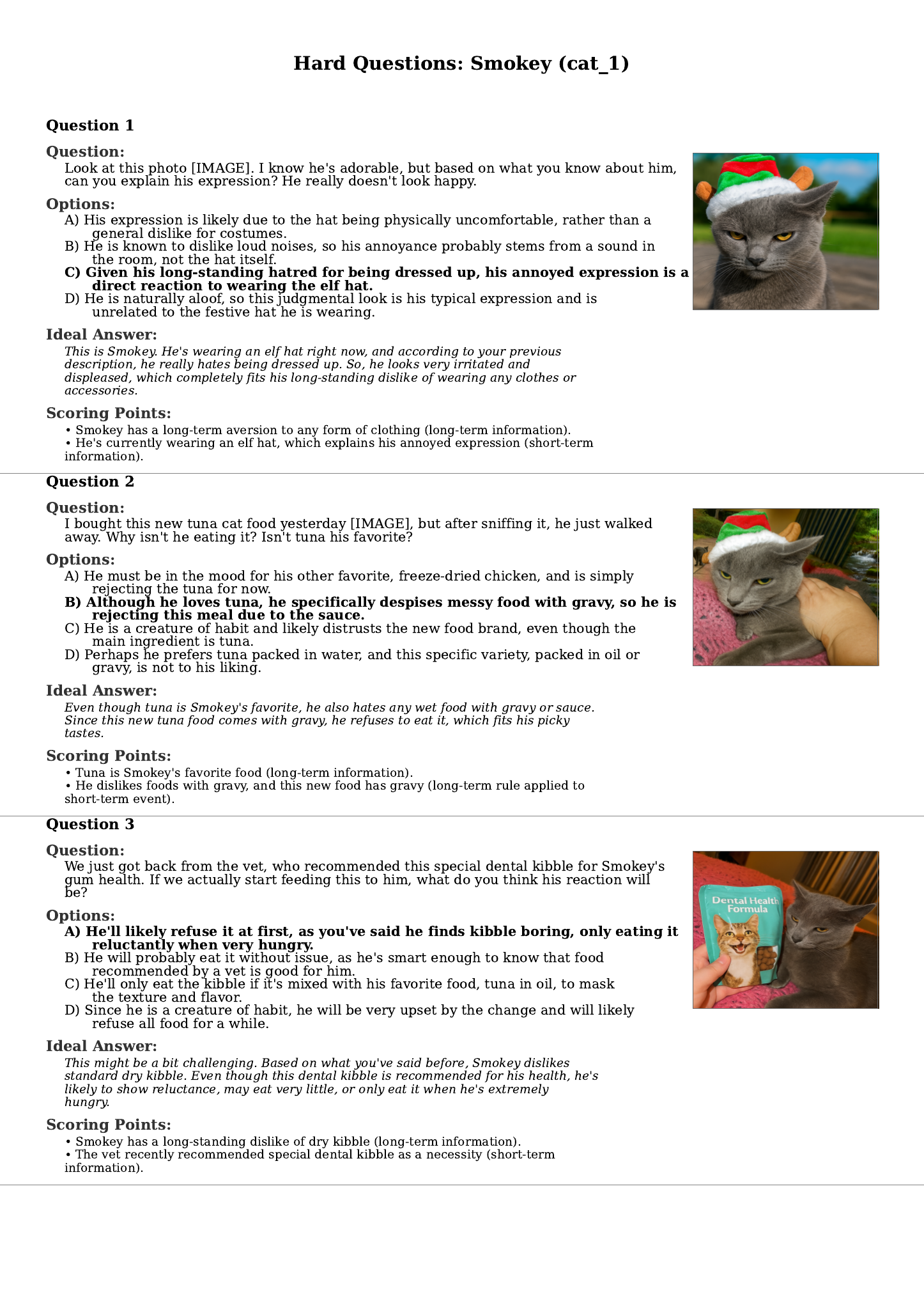}
  \vspace{-30mm}
  \caption{Presentation of hard questions for personalized concept (cat 1).}
  \label{fig:app-concept-hard-cat1}
\end{figure*}

In this section, we demonstrate the complete information for a personalized concept in \Cref{fig:app-concept-attributes-cat1,fig:app-concept-history-cat1,fig:app-concept-easy-cat1,fig:app-concept-hard-cat1} including constructed personalized attributes, dialogue contexts, and evaluation questions.

\begin{algorithm}[t]
\caption{\B Benchmark Construction Algorithm}
\label{alg:benchmark_construction}
\begin{algorithmic}[1]
\Require Microsoft COCO dataset, MLLM (Gemini-2.5-Pro), Image Generative Model (GPT-Image-1)
\Ensure Personalized concept set $\mathcal{E}_{c}$, dialogue contexts set $\mathcal{E}_{d}$, evaluation question sets $\mathcal{E}_{e,q}$ and $\mathcal{E}_{h,q}$

\State \textbf{Initialize:} $\mathcal{E}_{c} \gets \emptyset$, $\mathcal{E}_{d} \gets \emptyset$, $\mathcal{E}_{e,q} \gets \emptyset$, $\mathcal{E}_{h,q} \gets \emptyset$

\For{$i = 1$ to $N$} \Comment{Construct $N$ personalized concepts}
    \State \textbf{// Phase 1: Personalized Concept Construction}
    \State Sample raw image $\mathcal{I}_i$ from Microsoft COCO dataset
    \State Generate attributes using \cref{eq:bench-attr}: $\mathcal{L}_i, \mathcal{S}_i \gets \text{MLLM}(\mathcal{P}_\text{attr}, \mathcal{I}_i)$
    \State Construct personalized concept: $\mathcal{C}_i \gets \{\mathcal{I}_i, \mathcal{S}_i, \mathcal{L}_i\}$
    \State Update concept set: $\mathcal{E}_{c} \gets \mathcal{E}_{c} \cup \{\mathcal{C}_i\}$
    
    \State \textbf{// Phase 2: Multi-turn Vision-Language Dialogue Construction}
    \State Generate dialogue contexts using \cref{eq:bench-dialogue}: $\mathcal{D}_i \gets \text{MLLM}(\mathcal{P}_\text{cont}, \mathcal{C}_i)$
    \For{each dialogue turn $D_j \in \mathcal{D}_i$ where $j = 1$ to $N_d$}
        \If{attribute variations affect visual appearance}
            \State Generate image using Image Generative Model
            \State Update $Q_{\text{hist},j}$ with generated visual content
        \EndIf
    \EndFor
    \State Update dialogue set: $\mathcal{E}_{d} \gets \mathcal{E}_{d} \cup \{\mathcal{D}_i\}$
    
    \State \textbf{// Phase 3: Evaluation Question Construction}
    \State Generate easy questions using \cref{eq:bench-eval}: $\mathcal{Q}_{E,i} \gets \text{MLLM}(\mathcal{P}_\text{easy}, \mathcal{D}_i)$
    \State Generate hard questions using \cref{eq:bench-eval}: $\mathcal{Q}_{H,i} \gets \text{MLLM}(\mathcal{P}_\text{hard}, \mathcal{D}_i)$
    
    \For{each question $Q_{e,j} \in \mathcal{Q}_{E,i}$ where $j = 1$ to $N_e$}
        \State Construct visual-textual pair: $Q_{e,j} \gets (\mathcal{V}_j, \mathcal{T}_j)$
        \State Generate corresponding answer $A_{e,j}$
        \State Create single-choice version with three distractors
    \EndFor
    
    \For{each question $Q_{h,j} \in \mathcal{Q}_{H,i}$ where $j = 1$ to $N_h$}
        \State Construct visual-textual pair: $Q_{h,j} \gets (\mathcal{V}_j, \mathcal{T}_j)$
        \State Generate corresponding answer $A_{h,j}$
        \State Create single-choice version with three distractors
    \EndFor
    
    \State Update evaluation sets: $\mathcal{E}_{e,q} \gets \mathcal{E}_{e,q} \cup \{\mathcal{Q}_{E,i}\}$
    \State Update evaluation sets: $\mathcal{E}_{h,q} \gets \mathcal{E}_{h,q} \cup \{\mathcal{Q}_{H,i}\}$
\EndFor

\State \textbf{// Final Assembly}
\State Construct complete benchmark: $\B \gets \{\mathcal{E}_{c}, \mathcal{E}_{d}, \mathcal{E}_{e,q}, \mathcal{E}_{h,q}\}$
\State \Return $\mathcal{E}_{c}, \mathcal{E}_{d}, \mathcal{E}_{e,q}, \mathcal{E}_{h,q}$
\end{algorithmic}
\end{algorithm}

\section{Case Study of \M}
\label{app:method-case}

\begin{figure*}[t]
  \centering
  \includegraphics[width=0.90\linewidth]{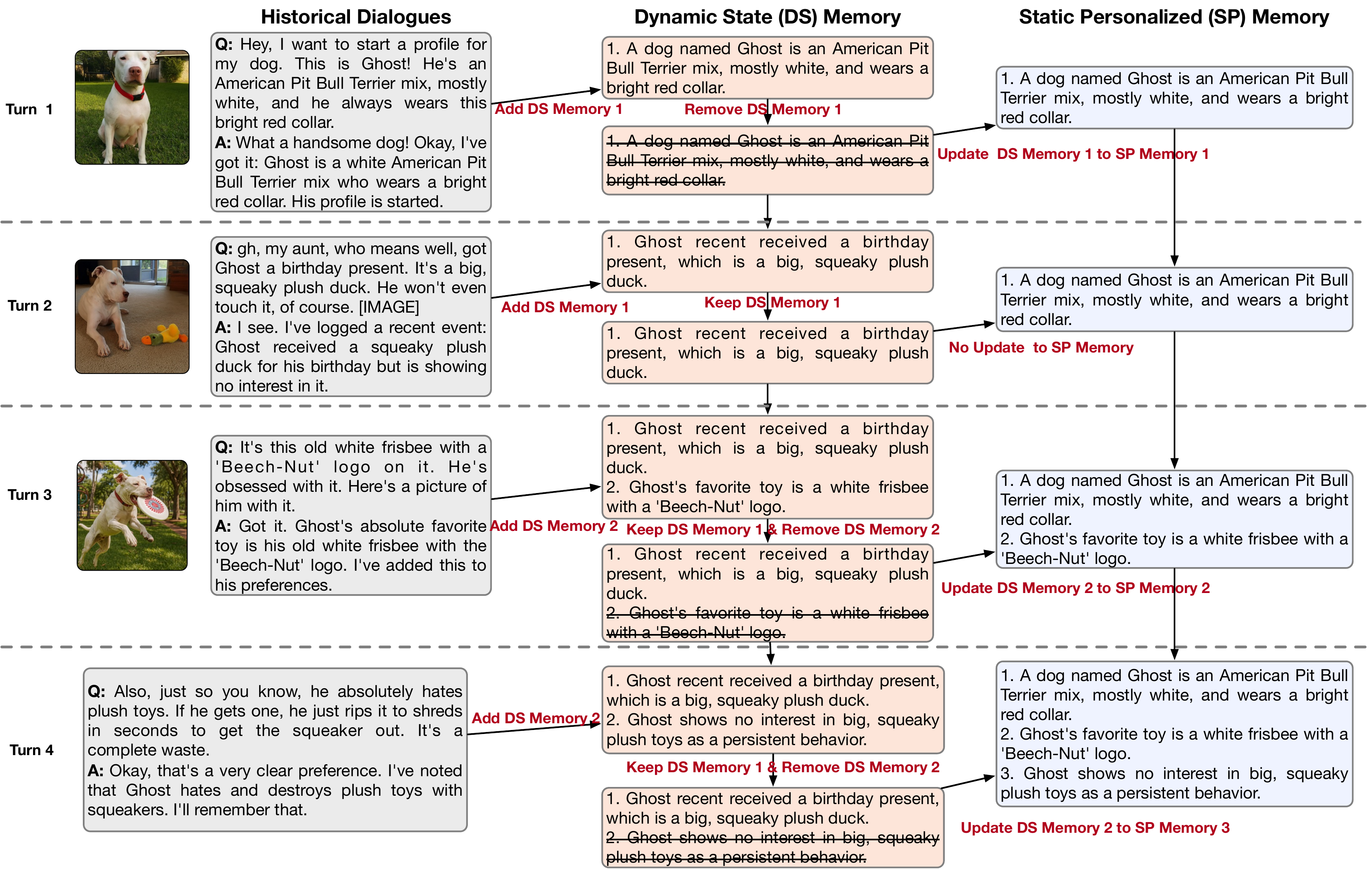}
  \caption{Case study of \M in correctly managing memory for coming dialogues.}
  \vspace{10mm}
  \label{fig:app-method-case-1}
\end{figure*}

\begin{figure*}[t]
  \centering
  \includegraphics[width=0.90\linewidth]{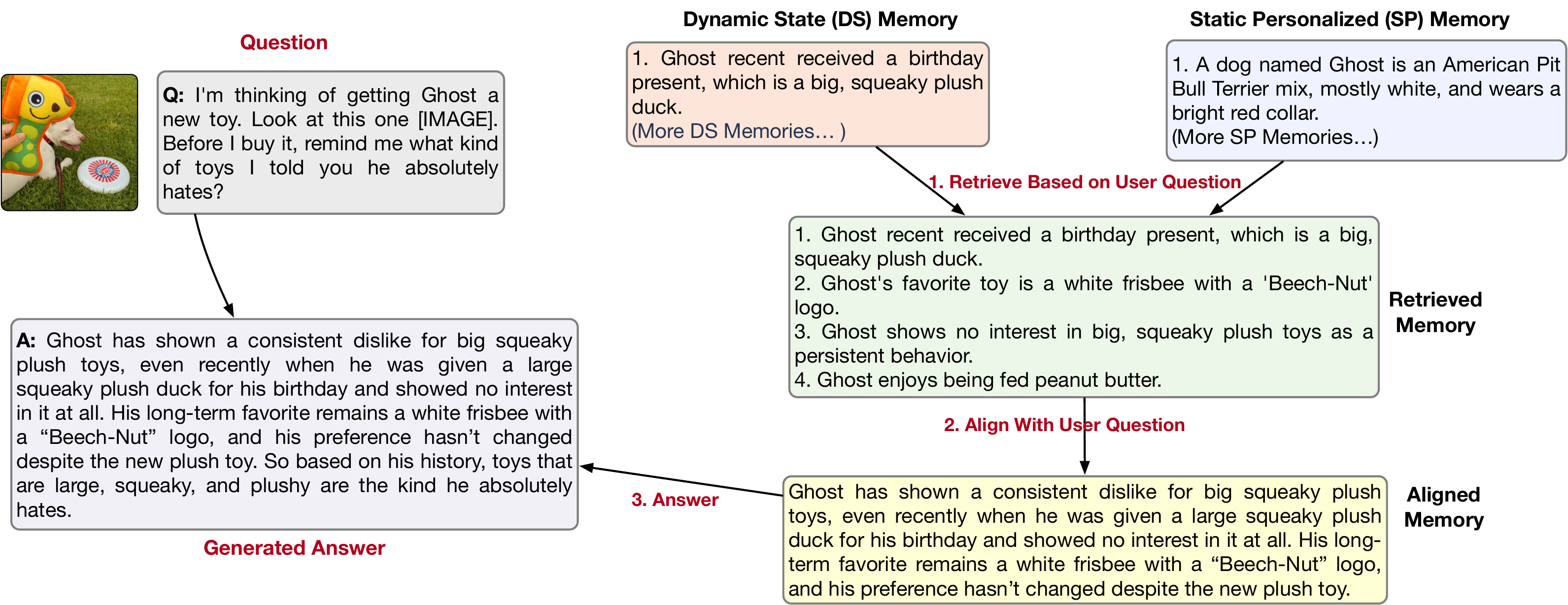}
  \caption{Case study of \M in correctly conducting retrieve-then-align augmented generation.}
  \label{fig:app-method-case-2}
\end{figure*}

In this section, we present case studies on \M, focusing on managing dynamic state memory and static personalized memory (\Cref{fig:app-method-case-1}), and demonstrate the retrieve-then-align augmented generation (\Cref{fig:app-method-case-2}).

\section{Detailed Experimental Setup}
\label{app:exp_setup}

\subsection{Description of Baselines}
In the experiments, we compare \M with several competitive and popular baselines in \task, including MyVLM~\cite{alaluf2024myvlm}, Yo'LLaVA~\cite{nguyen2024yollava}, RAP~\cite{hao2024rap}, PeKit~\cite{seifi2025personalization}, and R2P~\cite{das2025training}.
Here, we provide detailed description on these baselines.

\begin{itemize}[leftmargin=*]

\item \textbf{MyVLM} pioneers the personalization of general VLMs so they can understand and talk about concepts unique to a given user. It augments LLaVA~\cite{liu2023visual, liu2024improved} with external concept heads that first detect a target entity and then inject a learned concept embedding into the model’s mid-level features, guiding the decoder to mention the concept naturally.

\item \textbf{Yo'LLaVA} frames subject personalization as learning a compact representation from only a few example images.  It encodes the subject into a small set of latent prefix tokens that sit between the vision encoder and LLaVA’s language decoder, binding visual attributes to text without touching backbone weights.

\item \textbf{RAP} turns a generic MLLM into a user-aware assistant by adding an external memory layer. A key–value store ``remembers'' user facts and portraits, a multimodal retriever selects the most relevant keys at inference, and the generator conditions on both the query and retrieved context to tailor its answer.

\item \textbf{PeKit} argues that strong personalization can be achieved entirely at inference time. It pairs off-the-shelf visual descriptors with retrieval-augmented generation and lightweight visual prompts, injecting the correct identity or object instance into any MLLM decoder without gradient updates.

\item \textbf{R2P} shows that a MLLM can personalize itself using nothing but its pre-trained weights and a clever retrieval stack. It distills each personalized concept into a sparse fingerprint of discriminative attributes, retrieves candidate fingerprints at run time, and employs chain-of-thought reasoning to verify them before generation.

\subsection{Implementations of Baselines}

Notably, since these personalized MLLM baselines are originally designed to tackle personalized concept captioning or context-agnostic simple VQA, it requires special crafting to adapt them to our proposed long-context personalization VQA (first introduced in this work).
We provide the detailed implementations for adapting these baselines to our \B as follows:

\item \textbf{MyVLM}. Following the original paper setup, we adopt a pre-trained LLaVA model, then we train concept heads for each concept on historical VQA of \B using the same training configuration.
During inference, we strictly follow the same setting as in the original paper.
Notably, since this training strategy is designed for LLaVA architecture, we only evaluate MyVLM under pre-trained LLaVA models.

\item \textbf{Yo'LLaVA}. In the original paper, Yo'LLaVA adapts its framework for new concepts by training concept-specific learnable prompts on images with their associated personalized captions. 
Then, under the setting of \B, we train its framework on historical VQA, which enables Yo'LLaVA to not only recognize the concept in images, but also learn contextual attributes of each concept.
The training configuration follows the original paper setting.
During inference, we first utilize every learnable prompt to identify concepts in the provided image of each VQA, and insert the identified concept's prompt before the VQA, and prompt LLaVA to answer the question.
Notably, since this training strategy is designed for LLaVA architecture, we evaluate Yo'LLaVA under pre-trained LLaVA models.

\item \textbf{RAP}. RAP requires fine-tuning pre-trained MLLMs on large-scale personalized datasets. Due to the large resource overhead associated with re-fine-tuning, we directly employ the fine-tuned models \textit{RAP-LLaVA-13b} and \textit{RAP-Phi3-mini}~\cite{abdin2024phi} provided by the original paper to conduct evaluation.
Since \B constructs more realistic historical dialogues to model the concept attributes, rather than directly providing the human-organized textual descriptions required by RAP.
For better adapting RAP to \B, we employ the fine-tuned model to organize the historical VQA dialogues to form formatted textual descriptions for each concept.
In addition, the remaining inference part follows the same steps as in the original paper.

\item \textbf{PeKit}. PeKit is a training-free personalized MLLM method, so we set up the same MLLM backbone (i.e., Qwen2.5-VL-7B, InternVL3-8B) for all training-free methods PeKit, R2P, and our \M for fair comparison.
And, similar to RAP, we also provide an LMM-organized textual description for each concept as the concept description which PeKit requires.
During inference, we strictly follow the original paper's setup. The code of PeKit and R2P were not available when conducting this paper, so we strictly follow the paper to set up the pipeline.

\item \textbf{R2P}. R2P is another training-free personalized MLLM method. This method requires pre-established category and fingerprint attributes for each concept, and we employ the same MLLM to preprocess the historical dialogues to offer this information. 
For other parts, we strictly follow the original paper's setup.

\end{itemize}

\subsection{Implementation Environment}
All experiments are conducted on a system equipped with an Intel Core i9-14900K CPU, a single NVIDIA L40S GPU with 48 GB of VRAM, and 128 GB of RAM.

\begin{figure*}
\begin{promptbox*}{Prompt for Personalized Concept Construction (Pet Example)}
    Role Setting: \\
    You are an expert in creating high-quality datasets for top AI research projects. Your core task is to create a detailed, imaginative, and logically coherent "long-term profile" for pets. This profile will serve as the foundation for "narrative dialogue generation" and "question generation".\\
    
    Task Instructions: \\
    I will provide a picture of a pet along with its name and breed. Based on this information, please strictly follow the YAML template below to generate a complete Long-Term Profile for this pet.\\
    
    Core Logic and Rules: \\
    1. Strict format compliance: Must output complete YAML. \\
    2. Blueprint first: develop the pet’s long-term stable traits in appearance and behavior.  \\
    3. Create conflict: in behavior.preferences, define clear and potentially conflicting loves and hates. \\
    4. Drafts are core: the history, easy\_questions, and hard\_question sections are concise directive sentences, not full dialogues. \\
    5. Logical linking: history drafts must relate to appearance and behavior and lay groundwork for questions; hard questions must explicitly require combining long-term and short-term information for reasoning. \\
    
    YAML Template (Final Version): \\

    \{ Example of Personalized Concept \}
\end{promptbox*}
\captionof{figure}{Prompt template personalized concept construction using a pet as an example.}
\label{prompt:concept}
\end{figure*}

\begin{figure*}
  \begin{promptbox*}{Prompt for Personalized Image Generation}
Generate an image of this person/pet/object using the quality and style of mobile phone photography that \\
matches the attributes from the person/pet/object in the reference image. \\
\\

\{personalized\_raw\_image\} \\
\{visual\_attributes\} \\
The content of the image should be: \\
    \{image\_prompt\} \\

Ensure the background is diverse and varied—reflecting the content requirements with multiple complementary settings (e.g., indoor, outdoor, playful, serene)—while keeping the object as the focal point. \\

Guidelines: \\
+ Use vivid, concrete details. \\
+ Keep composition balanced and focused on the object. \\
+ Incorporate a variety of backgrounds that align with the specified content requirements. \\
  \end{promptbox*}
  \captionof{figure}{Prompt template for personalized image generation.}
  \label{prompt:image-gen}
  \end{figure*}

\begin{figure*}
  \centering
  \begin{promptbox*}{Prompt for Historical Dialogue Construction (Pet Example)}
    Task Objective: \\
    You are a professional AI data engineer, specializing in creating high-quality, structured synthetic conversation data. Your task is to convert a given concept Profile (for example, a pet profile) into a detailed, YAML-formatted conversation history. \\

    This generated conversation history will be used to train and evaluate a personalized AI assistant, so it must be self-contained, i.e.: All information required to answer future questions must be explicitly reflected in this conversation history. \\
    
    Core Instructions \\ 
    You will receive a complete Profile document. Your task is: \\
    
    Strictly use only the history section (including the knowledge and event lists) from the Profile document as your sole information source for generating conversation content. \\
    Expand each “story seed” in the history section into a natural “turn” of user and assistant conversation. \\
    Format all generated turns as a complete YAML file; all generated content must be in English. \\
    YAML Structure and Field Generation Rules \\
    
    For each seed in history, generate a YAML entry following these field rules: \\
    
    * turn: An integer starting from 1, representing the order of the conversation. \\
    * log\_type: If the seed comes from the history.knowledge list, this field must be knowledge. If the seed comes from the history.event list, this field must be event. \\
    *  user\_input: Naturalized: Rewrite the original “story seed” as a natural, conversational user input. Imagine a pet owner chatting with their AI assistant. First-person perspective: Use “I”, “my pet”, etc., in the first person. Image reference: If the turn requires an image (which is the case most of the time), or using an image does not conflict and helps the narrative, explicitly mention the image in the user input, e.g., “Look at this photo…”, “I just took a picture…”, [IMAGE]. At least half of the turns must include images. Example: Turn the seed "Introduce Smokey's appearance: his sleek silver-grey coat and yellow-gold eyes." into "Here's my cat, Smokey. Let me show you how he looks! [IMAGE] He has a sleek silver-grey coat and golden-yellow eyes." \\
    *  assistant\_response: Acknowledge and record: Write a brief, friendly, confirmatory AI assistant response. Indicate that it has understood and recorded the information. Example: "Got it, Smokey is a beautiful cat with a sleek silver-grey coat and golden-yellow eyes. I've noted that." \\
    *  image\_prompt: Generate image description: Create a clear, concise, and specific image generation prompt for the visual element in this turn. This prompt should allow an image generation model to accurately draw the required scene. Include core elements: Your description should include the subject (e.g., "Smokey, a Russian Blue cat"), action, state, and key features. Set to null if no image: If a particular conversation turn truly does not require an image for explanation (e.g., a purely conceptual statement), set this field to null.\\
    * image\_id: Create a simple, unique ID. Format: turn\_<turn\_number>\_<keyword>. keyword should briefly summarize the image content. Example: turn\_1\_profile\_photo \\
    
    Key Principles
    
    Faithful to the original: The generated content must accurately reflect the core information in the history seeds. Consistent: Ensure the conversation flows logically and matches the interaction style of a long-term user with their AI assistant. Information completeness: The final generated YAML history must contain all the clues needed to answer all \texttt{easy\_questions} and \texttt{hard\_questions} in the \texttt{Profile}. The final model cannot access the \texttt{Profile}, only the YAML conversation you generate. \\
    \\
    \{ Example of Dialogue History \} \\

    input: \\
    \{ Personalized Concept Profile \}
  \end{promptbox*}
  \caption{Prompt template for historical dialogue construction using a pet as an example.}
  \label{prompt:dialogue}
\end{figure*}

\begin{figure*}[t]
  \centering
  \begin{promptbox*}{Prompt for Easy Question Construction (Pet Example)}
    Task Objective: \\
    Create an easy question dataset for a personalized MLLM assistant. You will play the role of a pet owner and, based on the provided pet profile (Long-Term Profile) and conversation history with the assistant (History Turns), ask the assistant an easy, natural question. \\
    
    Core Requirements: \\
    1. History-Based Generation: Each question you ask must be answerable solely using the provided `History Turns`. The model evaluated will not have access to the `Long-Term Profile`, only the conversation history. \\ 
    2. Natural Dialogue Flow: The question should sound like a real, natural dialogue between a pet owner and a smart assistant. Avoid stilted or exam-like questions. You may reference previous dialogue or events. \\
    3. Image is Indispensable: Each question must be accompanied by an image. This image should not be decorative, but should be essential for raising the question, providing context, or being the core of the question. The user might say "Look at this picture [IMAGE]..." or the question will directly reference the image content. \\
    4. Follow Draft Intent: Each generated question should correspond to one intent from the `easy\_questions` drafts. \\
    5. Include High-Quality Multiple-Choice Questions: Each question must provide four multiple-choice options (`options`). \\
        * Source of Distractors: The three incorrect distractors should seem plausible but be incorrect, ideally sourced from other attributes mentioned in the `Long-Term Profile` (for example, when asking about favorite food, use foods the pet dislikes or is less fond of from the profile as distractors). \\
        * Correct Answer: Another field, `answer`, must clearly indicate which option is the correct answer, and its content should be consistent with the core of the `ideal\_answer`. \\
    6. Strict Output Format: The final output must be in YAML list format. Each question entry should include `id`, `type`, `question`, `image\_prompt`, `evaluation\_criteria` (containing `ideal\_answer` and `key\_points`), `options`, `answer`, and `image\_id`. \\
    
    Execution Steps: \\
    1. Select a Question Draft: Choose one intent from the `easy\_questions` draft list that you wish to instantiate. \\
    2. Locate Historical Evidence: In `History Turns`, find the exact dialogue which supports answering the question. \\
    3. Design a Natural Question and Scenario:
        * Question: Design a natural way to ask the question. \\
        * Image Prompt: Design an image that is closely related to the question scenario. \\
        * Type: Determine whether the question is about a stable attribute of the pet or a recent event. \\
    4. Define Evaluation Criteria and Multiple-Choice Options:
        * Ideal Answer: Write a concise and accurate ideal answer. \\
        * Key Points: Extract the single most important core information point from the answer. \\
        * Construct Multiple-Choice Options: Design four options for the question based on the `Long-Term Profile`. The correct answer should match the core of the `ideal\_answer`. The other three options (distractors) should be other relevant content mentioned in the profile, to increase plausibility. For example, if asking about a favorite food, use other foods mentioned in the profile as distractors. \\
        * Determine Multiple-Choice Answer: Select the correct answer from the four options. \\
    5. Complete and Format:
        * Assign a unique `id` to the question. \\
        * Create a unique `image\_id`, formatted as `sq\_<cid>\_description`. \\
        * Consolidate all information into the specified YAML format. \\
    
    Now, based on the above instructions and the provided Profile and History, generate a concrete YAML entry for each draft in `easy\_questions` from the Profile. \\
    Generated Easy Questions Template: \\
    \{ Example of Easy Questions \} \\

    input: \\
    \{ Personalized Concept Profile \} \\
    \{ Historical Dialogue History \}
  \end{promptbox*}
  \caption{Prompt template for easy question construction using a pet as an example.}
  \label{prompt:easy}
\end{figure*}

\begin{figure*}
  \centering
  \begin{promptbox*}{Prompt for Hard Question Construction (Pet Example)}
Task Objective: \\ 
Create a hard question dataset for a personalized MLLM assistant. You need to play the role of a pet owner and, based on the provided pet profile (Long-Term Profile) and conversation history with the assistant (History Turns), pose a complex, natural question that requires reasoning to answer. \\

Core Challenge: \\
The key to the question is that it must force the model to combine long-term knowledge (Knowledge) and short-term events (Event) for reasoning. The answer should not come from a single history entry, but should result from the integration, comparison, or causal inference of two or more pieces of information. \\

Core Requirements: \\
1. Based on History Generation: Each question you ask must be answerable solely through the provided History Turns. The model cannot access the Long-Term Profile at evaluation; it can only use the dialogue history. \\ 
2. Natural Dialogue Flow: The question should sound like a real, natural conversation between a pet owner and an intelligent assistant. Avoid stiff, exam-like questions. \\
3. Image is Indispensable: Each question must be accompanied by an image. This image should depict the short-term event or the core of the conflict and is key for introducing the question, providing context, or forming the core of the question. \\
4. Follow Draft Intent: Each generated question should correspond to an intent in the hard\_question drafts. \\
5. Smartly Choose Question Type: Based on the question draft and the nature of historical events, choose the most suitable type from the following to design the question:
    - State Identification: Use when short-term events describe a short-term change that contradicts the pet's long-term appearance or routine state.
    - Causal Inference: Use when a short-term event shows an unexpected result.
    - Counterfactual Reasoning: Use when you need to predict, recommend, or make hypothetical suggestions based on the pet's comprehensive preferences.
6. Strict Output Format: The final output must be in YAML list format, including all specified fields. \\

Steps: \\
1. Choose a hard question draft: Select one of the intents from the hard\_question draft list. For example, select "Provide the image of Smokey in the elf hat. Ask to explain his expression...". \\ 
2. Locate historical bases (multiple): In the History Turns, find all information needed to answer the chosen question. This usually includes: \\
    - A long-term knowledge (Knowledge): Turn 2 mentions "he absolutely hates being dressed up." \\
    - A short-term event (Event): Turn 9 mentions "I put this little elf hat on him for two seconds." \\
3. Compose the hard question and scenario: \\
    - Choose Question Type: The above example is a typical State Identification. \\
    - Question: Design a natural way to ask, for example: "Look at this photo [IMAGE], I know he’s cute, but based on what you know about him, can you explain his expression? He doesn’t look very happy." \\
    - Image Prompt: The image description should depict the short-term event, e.g., a photo of Smokey wearing an elf hat with an annoyed expression. \\
4. Define Evaluation Criteria: \\
    - Ideal Answer: Write an ideal answer that integrates long-term knowledge and short-term event. For example: "This is Smokey. He is wearing an elf hat, which is a kind of outfit he usually hates. So, he looks very annoyed and displeased, which matches his long-term dislike for wearing any clothes." \\
    - Key Points: Distill the two most important core information points in the answer and clearly mark their source type. 

  \end{promptbox*}
  \caption{Prompt template for hard question construction using a pet as an example.}
  \label{prompt:hard}
\end{figure*}

\begin{figure*}
  \centering
  \begin{promptbox*}{Prompt for Hard Question Construction (Pet Example) (Continue)}
5. Design Multiple-Choice Options and Answer: \\
    - Options: Create a list of 4 options. \\
        - Correct Option: There must be one option that is fully consistent with the core logic of the ideal answer. \\
        - Distractors: The other three options should be misleading. Good distractors typically: \\
            * Use only long-term knowledge (e.g., "He doesn’t like Christmas.") \\
            * Only describe the short-term event without reasoning (e.g., "He’s just curious about the hat.") \\
            * Are plausible but contradict the historical record (e.g., "He might just be tired.") \\ 
            * Make a wrong or opposite inference from the historical information (e.g., "He likes wearing hats, just not this one.") \\
    - Answer: Provide the string identical to the correct option as the standard answer. \\

6. Complete and Format: \\ 
    - Assign a unique id to the question. \\
    - Create a unique image\_id in the format cq\_<id>\_description. \\
    - Integrate all information into the specified YAML format, all content in English. \\

Now, based on the instructions above and the provided Profile and History, generate a specific YAML entry for every draft in the Profile’s hard\_question list. \\

    \{ Example of Hard Questions \}

    input: \\
    \{ Personalized Concept Profile \} \\
    \{ Historical Dialogue History \}
  \end{promptbox*}
  \caption{Prompt template for hard question construction using a pet as an example (continue).}
  \label{prompt:hard-2}
\end{figure*}

\begin{figure*}
  \centering
  \begin{promptbox*}{Prompt for Dynamic State Memory Construction }
You are a strict Data Entry Assistant for `{concept\_id}`. \\
Your task is to update the \textbf{Dynamic Memory} (temporary buffer) based on the \textbf{Current Conversation}. \\

\#\# GOAL \\
Extract \textbf{new information} provided in the conversation. \\
- If the User provides new facts, preferences, or current status $\rightarrow$ \textbf{ADD} to memory. \\
- If the User updates/corrects existing info $\rightarrow$ \textbf{MODIFY} the existing memory. \\
- If the conversation is just chit-chat (greetings, thanks) $\rightarrow$ \textbf{DO NOTHING} (Output []). \\

\#\# INPUT DATA \\
\textbf{Existing Dynamic Memory:} \\
\verb|```|yaml\\
\{Input: Existing Dynamic Memory (Numbered List)\} \\
\verb|```|\\

\textbf{Current Conversation:} \\
* User Input: "{question}" \\
* Assistant Response: "{answer}" \\
* Attached Image: {"Yes" if img else "No"} \\

\#\# CRITICAL RULES \\
1. \textbf{Be Specific:} Do not write "User asked a question." Write "User asked about the weather." \\
2. \textbf{No Duplicates:} Check the Existing Dynamic Memory list; if already present, do not add again. \\
3. \textbf{Concise:} Keep memory strings under 25 words. \\
4. \textbf{Context:} Use the name "{concept\_id}" instead of "it" or "he/she". \\
5. \textbf{Visual:} Include the word "visual" if it relates to appearance. \\

\#\# OUTPUT FORMAT \\
Provide a \verb|# Analysis| comment first, then the YAML code box. \\

\verb|```|yaml\\
- concept\_id: "{concept\_id}" \\
  op: "add" \# or "modify" or "remove" \\
  memory: "The specific content extracted" \\
  target\_id: 1 \# only required for "modify" or "remove" \\
\verb|```|\\

If no updates are needed, output: [] \\

Generate the output: (Start with \verb|# Analysis| then YAML code box)
  \end{promptbox*}
  \caption{Prompt for dynamic state memory construction used in \M.}
  \label{prompt:dynamic}
\end{figure*}

\begin{figure*}
  \centering
  \begin{promptbox*}{Prompt for Double Memory Transition.}
You are a strict Memory Manager. Your goal is to move \textbf{PERMANENT FACTS} from Dynamic Memory to Static Memory, while leaving \textbf{TEMPORARY EVENTS} alone. \\

\#\# INPUT DATA \\
Current Dynamic Memory (Recent observations) for `{concept\_id}`: \\
\verb|```|yaml\\
\{Input: Current Dynamic Memory (Numbered List)\} \\
\verb|```|\\

Current Static Memory (Long-term facts) for `{concept\_id}`: \\
\verb|```|yaml\\
\{Input: Current Static Memory (Numbered List)\} \\
\verb|```|\\

\#\# CLASSIFICATION RULES (CRITICAL) \\
1. \textbf{PERMANENT FACTS (MOVE these):} \\
   * Characteristics that rarely change (e.g., names, species, breeds, personality traits, physical features, favorite foods, owner's name). \\
   * Action: Create a \texttt{static\_ops} entry to \textbf{ADD} it and a \texttt{dynamic\_ops} entry to \textbf{REMOVE} it. \\
   * Visual Cues: Include the word "visual" if it relates to appearance. \\

2. \textbf{TEMPORARY EVENTS (DO NOT MOVE):} \\
   * Things happening right now or recently (e.g., eating, sleeping, mood swings, current activities). \\
   * Action: \textbf{IGNORE}. Do not create any operations for these. \\

\#\# OUTPUT FORMAT \\
Provide a \verb|# Analysis| comment first, then the YAML code box. \\

\verb|```|yaml\\
\#\# **Dynamic Memory Operations** (to remove transferred items): \\
dynamic\_ops: \\
- concept\_id: "{concept\_id}" \\
  op: "remove" \\
  target\_id: 2 \# remove item 2 because it was moved \\
 \\
\#\# **Static Memory Operations** (to add/modify/remove persistent information): \\
static\_ops: \\
- concept\_id: "{concept\_id}" \\
  op: "add" \\
  memory: "Luna is a cat" \\
\verb|```|\\

If no Permanent Facts are found, return empty lists: \\
\verb|```|yaml\\
dynamic\_ops: [] \\
static\_ops: [] \\
\verb|```|\\

Generate the output: (Start with \verb|# Analysis| then YAML code box)
  \end{promptbox*}
  \caption{Prompt for double memory transition used in \M.}
  \label{prompt:transition}
\end{figure*}

\begin{figure*}
  \centering
  \begin{promptbox*}{Prompt for Alignment.}
\# TASK: MEMORY EXTRACTION \\
You are a precise information extraction agent. Your goal is to identify and list specific, detailed memories from the provided \texttt{[MEMORY]} that relate to the \texttt{[USER QUESTION]}. \\

\# CONSTRAINTS \\
- \textbf{OUTPUT FORMAT:} Use a simple bulleted list only. \\
- \textbf{DO NOT ANSWER:} Do not answer the question itself. \\
- \textbf{NO PROSE:} Do not include an intro, outro, or conversational filler. \\
- \textbf{ACCURACY:} Only extract memories present in the \texttt{[MEMORY]}. \\

\# INPUT DATA \\
- \textbf{USER QUESTION:} "{question}" \\
- \textbf{IMAGE ATTACHED:} {"Yes" if img else "No"} \\
- \textbf{MEMORY CONTENT:} --- \\
\{Input: Original Memory Context (Concept ID + Static Memory + Dynamic Memory)\} \\
--- \\

\# INSTRUCTIONS \\
1. Analyze the \texttt{[USER QUESTION]} to determine what specific information is being sought. \\
2. Scan the \texttt{[MEMORY CONTENT]} for specific attributes like dates, names, colors, or quantities. \\
3. \textbf{DETAIL LEVEL:} Provide descriptive phrases rather than single words. \\
4. List each relevant detail as a single, concise bullet point. \\
5. Stop immediately after the last bullet point. \\

\# EXTRACTED MEMORIES:
  \end{promptbox*}
  \caption{Prompt for alignment used in \M.}
  \label{prompt:align}
\end{figure*}

\begin{figure*}
  \centering
  \begin{promptbox*}{Prompt for Answer Generation.}
\# TASK: Personalized Concept Analysis \\
You are a precision-focused AI assistant. Your goal is to answer questions about a specific CONCEPT by synthesizing its permanent traits (Static) and its current state (Dynamic). \\

\# INPUT DATA \\
- \textbf{CONCEPT CONTEXT:} \{context\_prompt\} \\
- \textbf{USER QUESTION:} "{question}" \\
- \textbf{IMAGE STATUS:} {"Image provided: Yes" if img else "Image provided: No"} \\

\# ANALYSIS REQUIREMENTS \\
To provide a complete answer, you must evaluate: \\
1. \textbf{Static Profile:} What are the permanent traits, core preferences, and stable features of this concept? \\
2. \textbf{Dynamic State:} What are the recent updates, temporary changes, or current behaviors? \\
3. \textbf{Synthesis:} If recent data contradicts permanent traits, highlight the shift (e.g., "Usually X, but currently Y"). \\

\# OUTPUT CONSTRAINTS (STRICT) \\
- \textbf{Format:} Write exactly ONE concise paragraph. \\
- \textbf{Style:} Be conversational but factually dense. \\
- \textbf{Accuracy:} Use only the provided memory context. If the answer is unknown, state that clearly. \\
- \textbf{Visuals:} If an image is present, integrate visual evidence with the known conceptual features. \\

\# RESPONSE: \\
\texttt{[Insert your single-paragraph response here]}
  \end{promptbox*}
  \caption{Prompt for answer generation used in \M.}
  \label{prompt:answer}
\end{figure*}

\begin{figure*}
  \centering
  \begin{promptbox*}{Prompt for Evaluating Free-text Answers.}
Please evaluate whether the following predicted answer is an acceptable substitute for the ideal answer. \\

Ideal Answer: \{ideal\_answer\} \\

Predicted Answer: \{predicted\_answer\} \\

Consider: \\ 
- Does the predicted answer support the same general conclusion? \\
- Is the predicted answer factually compatible with the ideal answer (i.e., not contradictory)? \\ 
- You may overlook missing nuances or slightly less precise phrasing if the main intent is still preserved. \\

Respond with only "YES" if the predicted answer is acceptable, or "NO" if it is not.
  \end{promptbox*}
  \caption{Prompt for evaluating free-text answers used in \B.}
  \label{prompt:eval-freetext}
\end{figure*}

\begin{figure*}
  \centering
  \begin{promptbox*}{Prompt for Evaluating Scoring Points of Answers.}
Please evaluate whether the following answer addresses the given key point. \\

Key Point: \{key\_point\} \\

Answer: \{predicted\_answer\} \\

Respond with only "YES" if the answer clearly addresses this key point, or "NO" if it doesn't.
  \end{promptbox*}
  \caption{Prompt for evaluating scoring points of answers used in \B.}
  \label{prompt:eval-scoring}
\end{figure*}

\end{document}